\documentclass{article}

    \usepackage{natbib}

\usepackage[english]{babel}
\usepackage[utf8]{inputenc} 
\usepackage[T1]{fontenc}    
\usepackage{hyperref}       
\usepackage{url}            
\usepackage{booktabs}       
\usepackage{amsfonts}       
\usepackage{nicefrac}       
\usepackage{microtype}      
\usepackage{wrapfig}
\usepackage{caption}
\usepackage{macros}
\usepackage{shortcuts}
\usepackage{xcolor}

\usepackage{amsmath,amsthm,amssymb,mathrsfs,bbm,mathabx}
\usepackage{algorithm,algpseudocode}

\usepackage{tikz}
\usetikzlibrary{arrows}

\usepackage[capitalise]{cleveref}
\DeclareMathOperator*{\argmin}{arg\,min}

\newtheorem{theorem}{Theorem}
\newtheorem{assumption}{Assumption}
\newtheorem{lemma}{Lemma}

\newtheorem{definition}{Definition}
\newtheorem{remark}{Remark}
\Crefname{assumption}{Assumption}{Assumptions}

\newcommand*\samethanks[1][\value{footnote}]{\footnotemark[#1]}

\newcommand{\Ind}{\mathbbm{1}}

\title{Policy Evaluation with Latent Confounders via Optimal Balance}

\author{%
  Andrew~Bennett\thanks{Alphabetical order.} \\
  Cornell University \\
  \texttt{awb222@cornell.edu} 
  \and
  Nathan~Kallus\samethanks\\
  Cornell University\\
  \texttt{kallus@cornell.edu} 
}
\date{}

\begin{document}

\maketitle

\begin{abstract}
Evaluating novel contextual bandit policies using logged data is crucial in applications where exploration is costly, such as medicine. But it usually relies on the assumption of no unobserved confounders, which is bound to fail in practice. We study the question of policy evaluation when we instead have proxies for the latent confounders and develop an importance weighting method that avoids fitting a latent outcome regression model. We show that unlike the unconfounded case no single set of weights can give unbiased evaluation for all outcome models, yet we propose a new algorithm that can still provably guarantee consistency by instead minimizing an adversarial balance objective. We further develop tractable algorithms for optimizing this objective and demonstrate empirically the power of our method when confounders are latent.
\end{abstract}


\section{Introduction}
\label{sec:intro}

Personalized intervention policies are of increasing importance in education \citep{mandel2014offline}, healthcare \citep{bertsimas2017personalized}, and public policy \citep{kube2019allocating}.
In many of these domains exploration is costly or otherwise prohibitive, and so it is crucial to evaluate new policies using existing observational data.
Usually, this relies on an assumption of no unobserved confounding (aka unconfoundedness or ignorability): that conditioned on observables, interventions are independent of idiosyncrasies that affect outcomes, so that counterfactuals can be reliably and correctly predicted. In particular, this enables the use of inverse propensity score (IPS) estimators of policy value \citep{beygelzimer2009offset,li2011unbiased,kallus2018policy} that eschew the need to actually fit outcome prediction models and doubly robust estimators that work even if such models are misspecified \citep{dudik2011doubly}.

In practice, however, it may be unlikely that we observe confounders exactly. Nonetheless, if we observe very many features they may serve as good proxies for the true confounders, which can enable an alternative route to identification \citep{louizos2017causal,kallus2018causal}. 
In particular, noisy observations of true confounders can serve as valid proxies.
For example, if intelligence is latent but affects both selection and outcome, we can instead use many noisy observations of intelligence such as school grades, IQ test, etc. Similarly, many medical measurements taken together can serve as proxies for underlying healthfulness.

In this paper, we study the problem of policy evaluation from observational data where we observe proxies instead of true confounders and we develop new weighting estimators based on optimizing balance in the latent confounders. Unlike the unconfounded setting where IPS weights ensure balance regardless of outcome model, we show that in this new setting there cannot exist any single of weights that ensure such unbiasedness regardless of outcome model. Instead, we develop an adversarial objective that bounds the conditional mean square error (CMSE) of any weighted estimator and, by appealing to game theoretic and empirical process arguments, we show that this objective can actually be driven to zero by a single set of weights. We therefore propose a novel policy evaluation method that minimizes this objective, thus provably ensuring consistent estimation in the face of latent confounders. We develop tractable algorithms for this optimization problem. Finally, we provide empirical evidence demonstrating our method's consistent evaluation compared to standard evaluation methods and its improved performance compared to using fitted latent outcome models.

\section{Problem}

\subsection{Setting and Assumptions}
\label{sec:setting}

We consider a contextual decision making setting with $m$ possible treatments (aka actions or interventions).
Each unit is associated with 
a set of potential outcomes $Y(1),\dots,Y(m)\in\Rl$ corresponding to the reward/loss for each treatment, an observed treatment $T\in\{1,\dots,m\}$, an observed outcome $Y=Y(T)$, true but latent confounders $Z\in\mathcal Z \subseteq \mathbb{R}^p$, and observed covariates $X\in\mathcal X \subseteq \mathbb{R}^q$.
Our data consists of iid observations $X_{i},T_{i},Y_{i}$ of $X,T,Y$.
Both the latent confounders and potential outcomes of unassigned treatments are unobserved.
Note that $Y_i=Y_i(T_i)$ encapsulates the assumptions of {consistency} between observed and potential outcomes and non-interference between units.

A \emph{policy} is a rule for assigning the probability of each treatment option given the observed covariates $X$. Given a policy $\pi$, we use the notation $\pi_t(x)$ to indicate the probability of assigning treatment $t$ when observed covariates are $x$. We define the \emph{value} of a policy, $\tau^{\pi}$, as the expected outcome that would be obtained from following the policy in the population. Formally:%
\begin{definition}[Policy Value]
$\tau^{\pi} = \mathbb{E}[\sum_{t=1}^m \pi_t(X) Y(t)]$.
\end{definition}
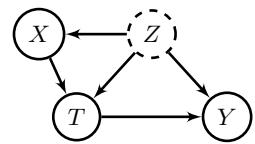
\begin{wrapfigure}{r}{0.225\textwidth}
\vspace{-2.75\baselineskip}
\begin{minipage}{0.225\textwidth}\small\centering%
\begin{tikzpicture}[%
    >=latex',
    circ/.style={draw, shape=circle, node distance=1cm, line width=1pt}]
    \node[circ] (X) at (-0.5,0.8) {$X$};
    \node[circ][draw,dashed] (U) at (1,0.8) {$Z$};
    \node[circ] (Y) at (2,-.3) {$Y$};
    \node[circ] (T) at (0,-.3) {$T$};
    \draw[line width=1pt,->] (T) -- (Y);
    \draw[line width=1pt,->] (U) -- (T);
    \draw[line width=1pt,->] (U) -- (Y);
    \draw[line width=1pt,->] (U) -- (X);
    \draw[line width=1pt,->] (X) -- (T);
\end{tikzpicture}
\end{minipage}%
\captionof{figure}{DAG representation of problem.}\label{fig:proxies}
\vspace{-\baselineskip}
\end{wrapfigure}

We encapsulate the assumption that $Z$ are sufficient for unconfoundedness and that $X$ is a proxy for $Z$ in the following assumption. \Cref{fig:proxies} provides a representation of this setting using a causal DAG \citep{pearl2000causality}.
Note importantly that we do \emph{not} assume ignorability given $X$. 
\begin{assumption}[$Z$ are true confounders]
\label{asm:ignorability}
For every $t\in\{1,\dots,m\}$, $Y(t)$ is independant of $(X,T)$, given $Z$.
\end{assumption}

We next define the average mean outcome given $Z$ and its conditional expectations given observables:
\begin{align*}
\mu_t(z) &= \mathbb{E}[Y(t) \mid Z=z],\\
\nu_t(x,t') &= \mathbb{E}[\mu_t(Z) \mid X=x, T=t']= \mathbb{E}[Y(t) \mid X=x, T=t'], \\
\rho_t(x) &= \mathbb{E}[\mu_t(Z) \mid X=x]= \mathbb{E}[Y(t) \mid X=x].
\end{align*}
We further define the propensity function and its conditional expectation given observables:
\begin{align*}
e_t(z) &= \Prb{T=t\mid Z=z},\\
\eta_t(x) &= P(T=t \mid X=x)=\Eb{e_t(Z)\mid X=x}.
\end{align*}

Finally, we denote by $\varphi(z; x,t)$ the conditional density of $Z$ given $X=x,T=t$. This density represents the latent variable model underlying the observables. For example, this can be a Gaussian mixture model, a PCA-type model as in \citep{kallus2018causal}, or a deep variational autoencoder as in \citep{louizos2017causal}. Because we focus on how one might use such a latent model rather than the estimation of this model, we just assume we have some oracle for calculating its values. (Note that for fair comparison, in experiments in \cref{sec:experiments}, we similarly let the outcome regression methods use this oracle.)

We further make the following regularity assumptions:
\begin{assumption}[Weak Overlap]
\label{asm:overlap}
$\Eb{e^{-2}_t(Z)}<\infty$
\end{assumption}

\begin{remark}
\label{rmk:overlap}
Given \cref{asm:overlap} it trivially follows that for every $t \in \{1,\ldots,m\}$, $x \in \mathcal X$, $z \in \mathcal Z$ that $e_t(z) > 0$ and $\eta_t(x) > 0$.
\end{remark}

\begin{assumption}[Bounded Variance]
\label{asm:variance}
The conditional variance of our potential outcomes given $X,T$ is bounded: $\mathbb{V}[Y(t) \mid X, T]\leq\sigma^2$.
\end{assumption}

\subsection{The Policy Evaluation Task}

The problem we consider is to estimate the policy value $\tau^\pi$ given a policy $\pi$ and data $X_{1:n},T_{1:n},Y_{1:n}$. One standard approach to this is the \emph{direct method} \citep{qian2011performance}, which given an estimate $\hat \rho_t$ of $\rho_t$ predicts the policy value as
\begin{equation}
\label{eq:direct}
\hat{\tau}^\pi_{\hat\rho} = \frac{1}{n} \sum_{i=1}^n \sum_{t=1}^m \pi_t(X_i) \hat{\rho}_t(X_i).
\end{equation}

However this method is known to be biased and doesn't generalize well \citep{beygelzimer2009offset}. Furthermore given \cref{asm:ignorability} $\hat \rho$ is not straightforward to estimate, since the mean value of $Y$ observed in our logged data given $X=x$ and $T=t$ is  $\nu_t(x,t)$ not $\rho_t(x)$, so fitting $\hat \rho$ would require controlling for the effects of the unobserved $Z$.

An alternative to this is to come up with weights $W_{1:n}=f_W(X_{1:n},T_{1:n})$ according to some function $f_W$ of the observed covariates and treatments, in order to re-weight the outcomes to look more like those that would be observed under $\pi$. Using these weights we can define the \emph{weighted} estimator
\begin{equation}\label{eq:weighted}
\hat{\tau}^{\pi}_W = \frac{1}{n} \sum_{i=1}^n W_i Y_i.
\end{equation}

This weighted estimator has the advantage that it does not require modeling the outcome distributions. Furthermore we could combine the weights $W_{1:n}$ with an outcome model $\hat \rho_t$ to calculate the \emph{doubly robust} estimator \citep{dudik2011doubly}, which is defined as
\begin{equation}
\label{eq:dr}
\hat{\tau}^{\pi}_{W,\hat{\rho}} = \frac{1}{n} \sum_{i=1}^n \sum_{t=1}^m \pi_t(X_i) \hat{\rho}_t(X_i) + \frac{1}{n} \sum_{i=1}^n W_i (Y_i - \hat{\rho}_{T_i}(X_i)).
\end{equation}
The doubly robust estimator is known to be consistent when either the weighted or direct estimator is consistent and can attain local efficiency \citep{chernozhukov2016double,robins2000robust,robins1994estimation,scharfstein1999adjusting}. 

Various approaches exist for coming up with weights for either the weighted or doubly robust estimators, which we discuss below. However none of these methods are applicable given \cref{asm:ignorability}, and so we develop a theory for weighting using proxy variables in \cref{sec:method}.


\subsection{Related Work}


One of the most standard approaches for policy evaluation is using the weighted or doubly robust estimator defined in \cref{eq:weighted,eq:dr}, using inverse propensity score (IPS) weights. These are given by $W_i = \pi_{T_i}(X_i) / e_{T_i}(Z_i)$ \citep{bottou2013counterfactual}, where $e_t$ are known or estimated logging probabilities. 
Since these weights can be extreme, both normalization \citep{austin2015moving,lunceford2004stratification,swaminathan2015self} and clipping \citep{elliott2008model,ionides2008truncated,swaminathan2015counterfactual} are often employed. In addition some other approaches include recursive partitioning \citep{kallus2016recursive}.
None of these methods are applicable to our setting however, since we do not know the true confounders $Z_{1:n}$.



An alternative to approaches based on fixed formulae for computing the importance weights is to compute weights that optimize an imbalance objective function \citep{athey2018approximate,kallus2017framework,kallus2018optimal}. For policy evaluation, \citet{kallus2018balanced} propose to choose weights that adversarially minimize the conditional mean squared error of policy evaluation in the worst case of possible mean outcome functions in some reproducing kernel Hilbert space (RKHS) class, by solving a linearly constraint quadratic program (LCQP). Our work follows a very similar style to this, however instead of using the true confounders we only assume access to proxies, and we prove our theory for more general families of functions. 

Finally there has been a long history of work in causal inference using proxies for true confounders \citep{wickens1972note,frost1979proxy}. As in our problem setup, much of this work is based on the model of using an identified latent variable model for the proxies \citep{wooldridge2009estimating,pearl2012measurement,kuroki2014measurement,miao2016identifying,edwards2015all}. Some recent work on this problem involves using techniques such as matrix completion \citep{kallus2018balanced} or variational autoencoders \citep{louizos2017causal} to infer confounders from the proxies. In addition, there is a variety of work that studies sufficient conditions for the identifiability of latent confounder models \citep{cai2008identifying,pearl2012measurement,miao2016identifying}. Our work is complementary to this line of research in that we assume access to an identified latent confounder model, but do not study how to identify such models. Furthermore our work is novel in combining proxy variable models with optimal balancing and applying it to finding importance weights for policy evaluation.






\section{Weight-Balancing Objectives}
\label{sec:method}

\subsection{Infeasibility of IPS-Style Unbiased Weighting}

If we had unconfoundedness given $X$ (\ie, $Y(t)\indep T\mid X$), the IPS weights $\pi_T(X)/\eta_T(X)$ are immediately gotten as the solution to making every term in the weighted sum \cref{eq:weighted} unbiased:
\begin{equation}
\label{eq:ips}
\mathbb{E}[W(X, T) \delta_{T_i t} Y(t)] = \mathbb{E}[\pi_t(X) Y(t)].
\end{equation}%
Notably the IPS weights do not depend on the outcome function.
However, without unconfoundedness given $X$ and given only \cref{asm:ignorability,asm:overlap,asm:variance}, this approach fails.
\begin{theorem}
\label{thm:ips}
If $W(x,t)$ satisfies \cref{eq:ips} then for any $t\in\{1,\dots,m\}$
\begin{equation}\label{eq:generalips} W(X,t) = \pi_t(X) \frac{\sum_{t' \in \mathcal{\tau}} \eta_{t'}(X) \nu_t(X,t') + \Omega_t(X)}{\eta_t(X) \nu_t(X,t)}, \end{equation}
for some $\Omega_t(x)$ such that $\E\Omega_t(X)=0\,\forall t$.
\end{theorem}
The proof of \cref{thm:ips} is given in \cref{apx:ips}. 

Note that \emph{if} we had unconfoundedness given $X$ then $\nu_t(x,t')=\nu_t(x,t)=\rho_t(x)$ so that
choosing $\Omega_t(X) = 0$ would recover the standard IPS weights. However, in our setting we generally have $\nu_t(x,t')\neq\nu_t(x,t)$, and so \cref{thm:ips} tells us that we cannot do unbiased IPS-style weighted evaluation without knowing the mean outcome functions $\nu_t(x,t')$. In particular, there exists no single weight function that is simultaneously unbiased for all outcome functions.


On the other hand, \cref{thm:ips} tells us that there do exist some weights that give unbiased and consistent policy evaluation via \cref{eq:weighted} or \cref{eq:dr}: we just may not be able to calculate them. 
The existence of such weights motivates our subsequent approach, which seeks weights that mimic these weights for a wide class of possible outcome functions.

\subsection{Adversarial Error Objective}

Over all weights that are functions of $X_{1:n},T_{1:n}$, the \emph{optimal} choice of weights for estimating $\tau^{\pi}$ via \cref{eq:weighted} would minimize the (unknown) conditional MSE (CMSE):
\begin{equation}\label{eq:cmsetau}\mathbb{E}[ (\hat{\tau}^{\pi}_W - \tau^{\pi})^2 \mid X_{1:n},T_{1:n} ].\end{equation} 
In particular, the weights in \cref{eq:generalips} achieve $O_p(1/n)$ control on this CMSE for many outcome functions, as long as the denominator is well behaved, which can be seen by applying concentration inequalities to \cref{eq:cmsetau}. However, as discussed above the outcome function is unknown and these weights are therefore practically infeasible.
Our aim is to find weights with similar near-optimal behavior but that do not depend on the particular unknown outcome function. To do this, we will find an upper bound for \cref{eq:cmsetau} that we can actually compute.

Let $f_{it} = W_i \delta_{T_i t} - \pi_t(X_i)$ and
$$
J(W,\mu) =  \left( \frac{1}{n} \sum_{i=1}^n \sum_{t=1}^m f_{it} \nu_t(X_i,T_i) \right)^2 + \frac{2\sigma^2}{n^2} \|W\|_2^2,
$$
where we embedded the dependence on $\mu$ inside $\nu_t(x,t')=\Eb{\mu_t(Z)\mid X=x,T=t'}$.

\begin{theorem}
\label{thm:jbound}
$\displaystyle\mathbb{E}[ (\hat{\tau}^{\pi}_W - \tau^{\pi})^2 \mid X_{1:n},T_{1:n} ]
\leq 2J(W,\mu)+O_p(1/n)
$.
\end{theorem}
We prove this in \cref{apx:jbound}.

Therefore, if we find weights that obtain $O_p(1/n)$ control on $J(W,\mu)$, we can ensure that we also have $O_p(1/n)$ control on $\mathbb{E}[ (\hat{\tau}^{\pi}_W - \tau^{\pi})^2 \mid X_{1:n},T_{1:n} ]$. Combined with the following result, which follows from \citep[Lemma~31]{kallus2016generalized}, this would give root-n consistent estimation.
\begin{lemma}
\label{lem:control}
If $\mathbb{E}[ (\hat{\tau}^{\pi}_W - \tau^{\pi})^2 \mid X_{1:n},T_{1:n} ]=O_p(1/n)$ then
$\hat{\tau}^{\pi}_W=\tau^{\pi}+O_p(1/\sqrt{n})$.
\end{lemma}

It remains to find weights that control $J(W,\mu)$. The key obstacle for this is that $\mu$ is unknown. Instead, we show how we can obtain weights that control $J(W,\mu)$ over a whole class of given functions $\mu$.

Suppose we are given a set $\mathcal{F}$ of functions mapping $\mathcal{Z}$ to $\mathbb{R}^m$, where each $\mu \in \mathcal{F}$ corresponds to a vector of mean outcome functions $\mu = (\mu_1, \ldots, \mu_m)$. Then, motivated by \cref{thm:jbound,lem:control}, we define our adversarial optimization problem as

\begin{equation}
\label{eq:optim}
\mathcal{W}^* = \argmin_{W \in \mathcal{W}} \sup_{\mu \in \mathcal{F}} J(W, \mu).
\end{equation}

One question the reader might ask at this point is why not solve the above optimization problem by ignoring the hidden confounders and directly balancing the conditional mean outcome functions $\nu_t(x,t')$. The problem is that this would be impossible to do over any kind of generic flexible function space, since we have no data corresponding to terms in the form $\nu_t(x,t')$ when $t \neq t'$, so this is akin to an overlap problem. Conversely, if we were to ignore the conditioning on $t$ and balance against functions of the form $\nu_t(x) = \nu_t(x,t)$ this would be inadequate, as we couldn't hope for such a space to cover the true $\mu$ since we don't assume ignorability given $Z$.

In light of these limitations, we can view what we are doing in optimizing \cref{eq:optim} using an identified model $\varphi(z;x,t)$ as implicitly balancing some controlled space of functions $\nu_t(x,t')$ that do not have this overlap issue between different $t$ values. The following lemma makes this explicit, as it implies that that the terms $\nu_t(x,t')$ are all mutually bounded by each other for fixed $x$ and $t$:

\begin{lemma}
\label{lem:gbound}
Assuming $\|\mu_t\|_\infty\leq b$, under \cref{asm:overlap}, for all $x \in \mathcal{X}$, and $t,t',t'' \in \{1,\ldots,m\}$ we have
\[ |\nu_t(x,t'')| \leq \frac{\eta_{t'}(x)}{\eta_{t''}(x)} \sqrt{8b
\Eb{e^{-2}_t(Z) \mid X=x, T=t'} |\nu_t(x,t')|}. \]
\end{lemma}

We prove this in \cref{apx:gbound}.

\subsection{Consistency of Adversarially Optimal Estimator}

Now we analyze the consistency of our weighted estimator based on \cref{eq:optim}. Given \cref{lem:control}, all we need to justify to prove consistency is that $\mu \in \mathcal{F}$ and that $\inf_W \sup_{\mu \in \mathcal{F}} J(W,\mu) \in O_p(\frac{1}{n})$. Define $\mathcal{F}_t$ as the space of all functions for treatment level $t$ allowed by $\mathcal{F}$. That is $\mathcal{F}_t = \{\mu_t : \exists (\mu_1', \ldots, \mu_m') \in \mathcal{F} \ \text{with} \ \mu_t' = \mu_t\}$. We will use the following assumptions about $\mathcal{F}$ to prove control of $J$:

\begin{assumption}[Normed]
\label{asm:norm}
For each $t \in \{1,\ldots,m\}$ there exists a norm $\|\cdot\|_t$ on $\op{span}(\mathcal{F}_t)$, and there exists a norm $\|\cdot\|$ on $\op{span}(\mathcal{F})$ which is defined given some $\mathbb{R}^m$ norm as $\|\mu\|~=~\|(\|\mu_1\|_1,\ldots,\|\mu_m\|_m)\|$.
\end{assumption}

\begin{assumption}[Absolutely Star Shaped]
\label{asm:symmetry} 
For every $\mu \in \mathcal{F}$ and $\abs{\lambda} \leq 1$, we have $\lambda \mu \in \mathcal{F}$.
\end{assumption}

\begin{assumption}[Convex Compact]
\label{asm:compact}
$\mathcal{F}$ is convex and compact 
\end{assumption}

\begin{assumption}[Square Integrable]
\label{asm:ltwo}
For each $t \in \{1,\ldots,m\}$ the space $\mathcal F_t$ is a subset of $\ltwo(\mathcal Z)$, and its norm dominates the $\ltwo$ norm (i.e., $\inf_{\mu_t \in \mathcal F_t} \|\mu_t\| / \|\mu_t\|_{\ltwo} > 0$).
\end{assumption}

\begin{assumption}[Nondegeneracy]
\label{asm:nondegen} 
Define $\mathcal B(\gamma) = \{\mu \in \op{span}(\mathcal F) : \| \mu \| \leq \gamma\}$. Then we have $\mathcal B(\gamma) \subseteq \mathcal F$ for some $\gamma > 0$.
\end{assumption}

\begin{assumption}[Boundedness]
\label{asm:bounded}
$\sup_{\mu \in \mathcal F} \|\mu\|_{\infty} < \infty$.
\end{assumption}

\begin{definition}[Rademacher Complexity]
$\mathcal R_n(\mathcal F) = \mathbb{E}[\sup_{f \in \mathcal F} \frac{1}{n} \sum_{i=1}^n \epsilon_i f(Z_i)]$, where $\epsilon_i$ are iid Rademacher random variables.
\end{definition}

\begin{assumption}[Complexity]
\label{asm:complexity} 
For each $t \in \{1\ldots,m\}$ we have $\mathcal R_n(\mathcal F_t) = o(1)$.
\end{assumption}

These assumptions are satisfied for many commonly-used families of functions, such as RKHS spaces and families of neural networks. We shall prove this claim for RKHS spaces in \cref{sec:algo}.

In order to justify that we can control $J$, first we will show that these assumptions allow us to reverse the order of minimization and maximization in our optimization problem. This means we can reduce problem to finding weights to control any particular $\mu$ rather than controlling all of $\mathcal{F}$.

\begin{lemma}
\label{lem:minimax}
Let $B(W, \mu) = \frac{1}{n} \sum_{i=1}^n \sum_{t=1}^m f_{it} \nu_t(X_i,T_i)$. Then under \cref{asm:symmetry,asm:compact,asm:ltwo} for every $M>0$ we have the bound 
\[ \min_W \sup_{\mu \in \mathcal{F}} J(W, \mu) \leq \sup_{\mu \in \mathcal F} \min_{\|W\|_2 \leq M} B(W, \mu)^2 + \frac{\sigma^2}{n^2}M^2. \]
\end{lemma}

We prove this in \cref{apx:minimax}.

Next, we note that \cref{lem:minimax} means that we can choose of weights given $\mu$ to set $B(W,\mu) = 0$, and therefore we have our desired control as long as we can justify that these weights have controlled euclidean norm. Using this strategy and optimizing for the weights of this kind with minimum euclidean norm, we are able to prove the following:

\begin{lemma}
\label{lem:jconvergence}
Under \cref{asm:norm,asm:symmetry,asm:compact,asm:ltwo,asm:nondegen,asm:bounded,asm:complexity} we have $\inf_W \sup_{\mu \in \mathcal{F}} J(W, \mu) = O_p(1/n)$.
\end{lemma}

We prove this in \cref{apx:jconvergence}. This is the key lemma in proving our main consistency theorem:

\begin{theorem}
\label{thm:consistency}
Under \cref{asm:norm,asm:symmetry,asm:compact,asm:ltwo,asm:nondegen,asm:bounded,asm:complexity} and assuming that $\mu \in \mathcal F$ we have $\hat{\tau}^{\pi}_{W^*} = \tau^{\pi} + O_p(1/\sqrt{n})$.
\end{theorem}

This theorem follows immediately from our previous results, since $\mu \in \mathcal F$ and \cref{lem:jconvergence} imply that $J(W^*, \mu) = O_p(1/n)$. This combined with \cref{thm:jbound} imply that $\mathbb{E}[ (\hat{\tau}^{\pi}_{W^*} - \tau^{\pi})^2 \mid X_{1:n},T_{1:n} ] = O_p(1/n)$, which in turn combined with \cref{lem:control} gives us our result.

\section{Algorithms for Optimal Kernel Balancing}
\label{sec:algo}

\subsection{Kernel Function Class}

We now provide an algorithm for optimal balancing when our function class consists of vectors of RKHS functions. Formally, given a kernel $K$ and corresponding RKHS norm $\|\cdot\|_K$, we define the space $\fk$ as follows:

\begin{definition}[Kernel Class]
\label{def:kernel-class}
$\fk = \{\mu : ||\mu|| \leq 1\}$, where $||(\mu_1,\ldots,\mu_m)|| = \sqrt{\sum_{t=1}^m ||\mu_t||_K^2}$.
\end{definition}

\begin{theorem}
\label{thm:fk}
Assuming $K$ is a Mercer kernel \citep{zhou2002covering} and is bounded, $\fk$ satisfies \cref{asm:norm,asm:symmetry,asm:compact,asm:ltwo,asm:nondegen,asm:bounded,asm:complexity}.
\end{theorem}

We prove this in \cref{apx:fk}.

We can remark that the commonly used Gaussian kernel is both Mercer and bounded, so it satisfies the conditions of \cref{thm:fk}. Given this, and assuming that $\fk$ covers the real mean outcome function $\mu$, we can apply \cref{thm:consistency} to see that solving \cref{eq:optim} using $\fk$ gives consistent evaluation.

We can note that the $\fk$ having maximum norm 1 is without loss of generality, because if we wanted the maximum norm to instead be $\gamma$ we could replace the $\Sigma$ matrix by $\Gamma = \frac{1}{\gamma} \Sigma$ in our objective function, resulting in an equivalent re-scaled optimization problem. To make this explicit, we will replace the $\Sigma$ matrix in the objective with $\Gamma$ in this section, where it is assumed that $\Gamma$ is freely chosen as a hyperparameter.

\subsection{Kernel Balancing Algorithm}
\label{sec:qpalgo}

In order to optimize \cref{eq:optim} over a class of kernel functions as defined by \cref{def:kernel-class}, we can observe that the definition of $J(W, \mu)$ looks very similar to the adversarial objective of \citet{kallus2018balanced}, except that we have $\nu_t(X_i,T_i)$ terms instead of $\mu_t(X_i)$ terms. This motivates the idea that, given our identified posterior model $\varphi(z;x,t)$, we may be able to employ a similar quadratic programming (QP)-based approach. The following theorem makes this explicit, by defining a QP objective for $W$ that we can approximate by sampling from $\varphi$:

\begin{theorem}
\label{thm:continuous-algorithm}
Define $Q_{ij} = \mathbb{E}[K(Z_i, Z_j')]$, $G_{ij} = \frac{1}{n^2} (Q_{ij} \delta_{T_i T_j} + \Gamma_{ij})$, and $a_i = \frac{2}{n^2} \sum_{j=1}^n Q_{ij} \pi_{T_j}(X_i)$, where for each i $Z_i$ and $Z_i'$ are iid shadow variables. Then for some $c$ that is constant in $W$ we have the identity
\[ \sup_{\mu \in \fk} J(W, \mu) = W^T G W - a^T W + c.\]
\end{theorem}

We prove this in \cref{apx:continuous-algorithm}.

Given this our balancing algorithm is natural and straightforward, and is summarized by \cref{algo:continuous}. Note that we provide an optional weight space constraint $\mathcal{W}$ in this algorithm, since standard weighted estimator approaches for policy evaluation regularize by forcing constraints such as $W \in n \Delta^n$. Under this kind of constraint our unconstrained QP becomes a LCQP. However that our theory does not support this constraint, and that we find it hurts performance in practice, especially when $\Gamma$ is large, so we do not use this constraint in our experiments.



\begin{algorithm}
  \caption{Optimal Kernel Balancing}\label{algo:continuous}
  \begin{algorithmic}[1]
  \Require \parbox[t]{0.9\textwidth}{
     Data $(X_{1:n}, T_{1:n})$, policy $\pi$, kernel function $K$, posterior density $\varphi$, regularization matrix $\Gamma$, number samples $B$, optional weight space $\mathcal{W}$ (defaults to $\mathbb{R}^n$ if not provided)
  }
  \Ensure Optimal balancing weights $W_{1:n}$
      \For {$i \in \{1, \ldots, n\}$}
        \State \parbox[t]{0.9\textwidth}{\textbf{Sample Data.} Draw $B$ data points $Z_i^b$ from the posterior $\varphi(\cdot \ ;X_i,T_i)$}
      \EndFor
      \State \textbf{Estimate Q.} Calculate $Q_{ij} = \frac{1}{B^2} \sum_{b=1}^B \sum_{c=1}^B K(Z_i^b, Z_i^c)$
      \State \textbf{Calculate QP Inputs.} Calculate $G_{ij} = Q_{ij} \delta_{T_i T_j} + \Gamma_{ij}$, and $a_i = 2 \sum_{j=1}^n Q_{ij} \pi_{T_j}(X_i)$,
      \State \textbf{Solve Quadratic Program.} Calculate $W = \argmin_{W \in \mathcal{W}} W^T G W - a^T W$
  \end{algorithmic}
\end{algorithm}


\section{Experiments}
\label{sec:experiments}

\subsection{Experimental Setup}
\label{sec:exp-setup}

We now present a brief set of experiments to explore our methodology. The aim of these experiments is to be a proof of concept of our theory. We seek to show that given an identified posterior model $\varphi$ policy as discussed in \cref{sec:setting}, evaluation using the weights defined by \cref{eq:optim} can give unbiased policy evaluation even in the face of sufficiently strong confounding where standard benchmark approaches that rely on ignorability given $X$ fail. We experiment with the following generalized linear model-style scenario:

\begin{center}
\begin{tabular}{lll}
$Z \sim \mathcal{N}(0,1)$ \qquad & \qquad $X \sim \mathcal{N}(\alpha^T Z + \alpha_0, \sigma^2_X)$ \qquad & \qquad $P_T = \beta^T Z + \beta_0$ \\
$T \sim \op{softmax}(P_T)$ \qquad & \qquad $W(t) \sim \mathcal{N}(\zeta(t)^T Z + \zeta_0(t), \sigma^2_Y)$ \qquad & \qquad $Y(t) = g(W(t))$ 
\end{tabular}
\end{center}

In our experiments $Z$ is $1$-dimensional, $X$ is $10$-dimensional, and we have two possible treatment levels ($m=2$). We experiment with a parametric policy $\pi$ and multiple link functions $g$ as follows:
$$\pi_t(X) = \frac{\exp(\psi_t^T X)}{\exp(\psi_1^T X) + \exp(\psi_2^T X)}$$
$$
\text{\textbf{step}: } g(w) = 3 \Ind_{\{w \geq 0\}} - 6 \qquad
\text{\textbf{exp}: } g(w) = \exp(w) \qquad
\text{\textbf{cubic}: } g(w) = w^3 \qquad
\text{\textbf{linear}: } g(w) = w 
$$

We experiment with the following methods in this evaluation:
\begin{enumerate}
\item \textbf{OptZ} Our method, using $\Gamma = \gamma \op{Identity}(n)$ for $\gamma \in \{0.001, 0.2, 1.0, 5.0\}$.
\item \textbf{IPS} IPS weights based on $X$ using estimated $\hat\eta_t$.
\item \textbf{OptX} The optimal weighting method of \citet{kallus2018balanced} with same values of $\Gamma$ as our method.
\item \textbf{DirX} Direct method by fitting $\hat \rho_t(x)$ incorrectly assuming ignorability given $X$.
\item \textbf{DirZ} Direct method by first fitting $\hat\mu_t$ using posterior samples from $\varphi$, then using the estimate $\hat \rho_t(x) = (1/D) \sum_{i=1}^D \hat\mu_t(z_i')$, where $z_i'$ are sampled from $\varphi(\cdot;x,t)$.
\item \textbf{D:W} Doubly robust estimation using direct estimator $\textbf{D}$ and weighted estimator $\textbf{W}$.
\end{enumerate}

Finally we detail all choices for scenario parameters in \cref{apx:exp-scenario}, and provide implementation details of our methods in \cref{apx:exp-benchmark}.\footnote{Code available online at \url{https://github.com/CausalML/LatentConfounderBalancing}.}

\subsection{Results}
We display results for our experiments using the \textbf{step} link function in \cref{tab:results-ours,tab:results-benchmarks}. For each of $n \in \{200, 500, 1000, 2000\}$ we estimate the RMSE of policy evaluation using each method, as well as doubly robust evaluation using our best performing weights, by averaging over 64 runs. In addition, in \cref{tab:results-ours-bias,tab:results-benchmarks-bias} we display the estimated bias from the evaluations. It is clear that the naive methods that assume ignorability given $X$ all hit a performance ceiling, where bias converges to some non-zero value. In particular for $\text{IPS}$ we separately ran it on up to one million data points and found that the bias converged to $0.418\pm0.001$. One the other hand, for our method it appears like we have consistency. This is particularly evident when we look at \cref{tab:results-ours-bias}, as bias seems to be approximately converging to zero with vanishing variance. We can also observe that doubly robust estimation using either direct method does not appear to improve performance.

It is noteworthy that the \textbf{DirZ} benchmark method fails horribly, despite being a correctly specified regression estimate. From our experience we observed that it is difficult to train the $\mu_t$ functions accurately if there is a high amount of overlap in the $\varphi(\cdot;x,t)$ posteriors for fixed $t$. Therefore we postulate that in highly confounded settings this benchmark inherently difficult to train using a finite number of samples from $\varphi(\cdot;x,t)$, and the result seems to collapse to degenerate solutions.

Finally we note that we observed similar trends to this using our other link functions, and other doubly robust estimators. We present more extensive tables of results in \cref{apx:exp-results}. In addition we present some results there on the negative impact on our method's performance using the constraint $W \in n\Delta^n$, as mentioned in \cref{sec:qpalgo}.

\begin{table}
\begin{center}
\footnotesize
\begin{tabular}{ccccccc}
\hline
n & $\textbf{OptZ}_{0.001}$ & $\textbf{OptZ}_{0.2}$ & $\textbf{OptZ}_{1.0}$ & $\textbf{OptZ}_{5.0}$ & $\textbf{DirX:OptZ}_{0.001}$ & $\textbf{DirZ:OptZ}_{0.001}$\\
\hline
200 & $.39\pm.07$ & $.24\pm.02$ & $.36\pm.02$ & $.81\pm.02$ & $.57\pm.06$ & $.41\pm.07$\\
500 & $.19\pm.02$ & $.18\pm.02$ & $.23\pm.02$ & $.49\pm.02$ & $.55\pm.03$ & $.20\pm.02$\\
1000 & $.11\pm.01$ & $.11\pm.01$ & $.13\pm.01$ & $.27\pm.01$ & $.49\pm.02$ & $.11\pm.01$\\
2000 & $.08\pm.01$ & $.08\pm.01$ & $.09\pm.01$ & $.17\pm.01$ & $.48\pm.01$ & $.08\pm.01$\\
\hline
\end{tabular}
\end{center}
\caption{Convergence of RMSE for for policy evaluation using our weights.}
\label{tab:results-ours}
%
\begin{center}
\footnotesize
\begin{tabular}{ccccccccccccc}
\hline
n & $\textbf{IPS}$ & $\textbf{OptX}_{0.001}$ & $\textbf{OptX}_{0.2}$ & $\textbf{OptX}_{1.0}$ & $\textbf{OptX}_{5.0}$ & $\textbf{DirX}$ & $\textbf{DirZ}$\\
\hline
200 & $.47\pm.03$ & $2.0\pm.03$ & $2.1\pm.03$ & $2.3\pm.02$ & $2.5\pm.02$ & $.52\pm.02$ & $2.6\pm.02$\\
500 & $.48\pm.03$ & $2.0\pm.02$ & $2.1\pm.02$ & $2.3\pm.02$ & $2.6\pm.02$ & $.48\pm.02$ & $2.6\pm.01$\\
1000 & $.39\pm.02$ & $2.0\pm.01$ & $2.1\pm.01$ & $2.3\pm.01$ & $2.5\pm.01$ & $.48\pm.02$ & $2.6\pm.01$\\
2000 & $.40\pm.01$ & $2.0\pm.01$ & $2.1\pm.01$ & $2.3\pm.01$ & $2.5\pm.01$ & $.45\pm.02$ & $2.6\pm.01$\\
\hline
\end{tabular}
\end{center}
\caption{Convergence of RMSE for benchmark methods.}
\label{tab:results-benchmarks}
%
\begin{center}
\footnotesize
\begin{tabular}{ccccccc}
\hline
n & $\textbf{OptZ}_{0.001}$ & $\textbf{OptZ}_{0.2}$ & $\textbf{OptZ}_{1.0}$ & $\textbf{OptZ}_{5.0}$ & $\textbf{DirX:OptZ}_{0.001}$ & $\textbf{DirZ:OptZ}_{0.001}$\\
\hline
200 & $.03\pm.39$ & $.11\pm.21$ & $.29\pm.21$ & $.78\pm.18$ & $.43\pm.38$ & $.05\pm.40$\\
500 & $.09\pm.17$ & $.10\pm.15$ & $.17\pm.16$ & $.47\pm.15$ & $.51\pm.19$ & $.10\pm.18$\\
1000 & $.02\pm.11$ & $.05\pm.09$ & $.08\pm.09$ & $.25\pm.09$ & $.47\pm.13$ & $.04\pm.11$\\
2000 & $.03\pm.07$ & $.05\pm.06$ & $.07\pm.07$ & $.16\pm.07$ & $.47\pm.09$ & $.03\pm.07$\\
\hline
\end{tabular}
\end{center}
\caption{Convergence of bias for policy evaluation using our weights.}
\label{tab:results-ours-bias}
%
\begin{center}
\footnotesize
\begin{tabular}{ccccccccccccc}
\hline
n & $\textbf{IPS}$ & $\textbf{OptX}_{0.001}$ & $\textbf{OptX}_{0.2}$ & $\textbf{OptX}_{1.0}$ & $\textbf{OptX}_{5.0}$ & $\textbf{DirX}$ & $\textbf{DirZ}$\\
\hline
200 & $.40\pm.25$ & $1.9\pm.21$ & $2.1\pm.20$ & $2.3\pm.19$ & $2.5\pm.18$ & $.49\pm.18$ & $2.6\pm.14$\\
500 & $.43\pm.21$ & $2.0\pm.16$ & $2.1\pm.15$ & $2.3\pm.14$ & $2.6\pm.13$ & $.45\pm.16$ & $2.6\pm.12$\\
1000 & $.37\pm.12$ & $2.0\pm.10$ & $2.1\pm.09$ & $2.3\pm.09$ & $2.5\pm.08$ & $.46\pm.15$ & $2.6\pm.11$\\
2000 & $.39\pm.10$ & $2.0\pm.08$ & $2.1\pm.07$ & $2.3\pm.07$ & $2.5\pm.07$ & $.42\pm.17$ & $2.6\pm.11$\\
\hline
\end{tabular}
\end{center}
\caption{Convergence of bias for benchmark methods.}
\label{tab:results-benchmarks-bias}
\end{table}


\section{Conclusion}

We presented theory for how to do optimal balancing for policy evaluation when we only have proxies for the true confounders, given an already identified model for the confounders, treatment, and proxies, but not for the outcomes. We provided an adversarial objective for selecting optimal weights given some class of mean outcome functions to be balanced, and proved that under mild conditions these optimal weights result in consistent policy evaluation. In addition, we presented a tractable algorithm for minimizing this objective when our function class is an RKHS, and we conducted a series of experiments to demonstrate that our method can achieve consistent evaluation even under sufficient levels of confounding where standard approaches fail.

For future work we note that the adversarial objective and theory presented here is fairly general, and could be used to develop new algorithms for balancing different function classes such as neural networks. An alternative direction would be to study how best to apply this methodology when an identified model is not already given.

\bibliography{ref}
\bibliographystyle{icml2018}


\appendix
\newpage

\section{Omitted Proofs}


\subsection{Proof of \cref{thm:ips}}
\label{apx:ips}

First we note that $W(X,T)Y - \sum_t \pi_t(X) Y(t)] = \sum_t (W(X,t) \delta_{Tt} - \pi_t(X)) Y(t)$. Then analyzing each summand separately, we can obtain:
\begin{align}
\mathbb{E}&[W(X,t) \delta_{Tt} Y(t) - \pi_t(X) Y(t)] \nonumber \\
&= \mathbb{E}[ \mathbb{E}[ W(X,t) \delta_{Tt} Y(t) - \pi_t(X) Y(t) \mid X] ] \nonumber \\
&= \mathbb{E}[ W(X,t) \mathbb{E}[\delta_{Tt} Y(t) \mid X] - \pi_t(X) \mathbb{E}[Y(t) \mid X] ] \nonumber \\
&= \mathbb{E}[ W(X,t) \mathbb{E}[\mathbb{E}[\delta_{Tt} Y(t) \mid X,T] \mid X] - \pi_t(X) \mathbb{E}[\mathbb{E}[Y(t) \mid X,Z] \mid X] ] \nonumber \\
&= \mathbb{E}[ W(X,t) \mathbb{E}[\delta_{Tt} \mathbb{E}[Y(t) \mid X,T=t] \mid X] - \pi_t(X) \mathbb{E}[\mu_t(Z) \mid X] ] \nonumber \\ 
&= \mathbb{E}[ W(X,t) \mathbb{E}[\delta_{Tt} \mid X] \mathbb{E}[\mu_t(Z) \mid X,T=t] - \pi_t(X) \mathbb{E}[\mu_t(Z) \mid X] ] \nonumber \\ 
&= \mathbb{E}[ W(X,t) \eta_t(X) \mathbb{E}[\mu_t(Z) \mid X,T=t] - \pi_t(X) \mathbb{E}[\mu_t(Z) \mid X] ] \nonumber \\ 
&= \mathbb{E}\left[ W(X,t) \eta_t(X) \nu_t(X,t) - \pi_t(X) \sum_{t'} \eta_{t'}(X) \nu_t(X,t') \right] \nonumber 
\end{align}

Therefore the solution to the equation $\mathbb{E}[W(X,t) \delta_{Tt} Y(t) - \pi_t(X) Y(t)] = 0$ is given by:
\begin{equation*}
W(X,t) \eta_t(X) \nu_t(X,t) - \pi_t(X) \sum_{t'} \eta_{t'}(X) \nu_t(X,t') = \Omega_t(X)
\end{equation*}
where $\Omega_t(X)$ is any arbitrary function of $X$ with mean zero. Solving this for $W$ gives
\begin{equation*}
W(X,t) = \frac{\pi_t(X) \sum_{t'} \eta_{t'}(X) \nu_t(X,t') + \Omega_t(X)}{\eta_t(X) \nu_t(X,t)},
\end{equation*}
and finally replacing $X$ with $x$ gives the required solution.

\subsection{Proof of \cref{thm:jbound}}
\label{apx:jbound}
Define 
$$\tau^{\pi}_{SAPE} = \frac{1}{n} \sum_i \sum_t \pi_t(X_i) \mu_t(Z_i).
$$
Using $(x+y)^2\leq2x^2+2y^2$, we have
\begin{align*}
\mathbb{E}[ (\hat{\tau}^{\pi}_W - \tau^{\pi})^2 \mid X_{1:n},T_{1:n} ]
\leq 2\mathbb{E}[ (\hat{\tau}^{\pi}_W - \tau^{\pi}_{SAPE})^2 \mid X_{1:n},T_{1:n} ]+2\mathbb{E}[ (\tau^{\pi}_{SAPE} - \tau^{\pi})^2 \mid X_{1:n},T_{1:n} ].
\end{align*}
Noting that $\tau^{\pi} = \mathbb{E}[\tau^{\pi}_{SAPE}]$, \Cref{asm:variance} implies that 
$\mathbb{E}[ (\tau^{\pi}_{SAPE} - \tau^{\pi})^2]=O(1/n)$.
Markov's inequality yields $\mathbb{E}[ (\tau^{\pi}_{SAPE} - \tau^{\pi})^2 \mid X_{1:n},T_{1:n} ]=O_p(1/n)$.

Let $CMSE(W,\mu)=\mathbb{E}[ (\hat{\tau}^{\pi}_W - \tau^{\pi}_{SAPE})^2 \mid X_{1:n},T_{1:n} ]$. We proceed to bound $CMSE$.
By iterating expectations we can obtain:
\begin{align}
CMSE(W, \mu) &= \mathbb{E}[ ( \hat{\tau}^{\pi}_W - \tau^{\pi}_{SAPE} )^2 \mid X, T ] \nonumber \\
&= \mathbb{E}[ \mathbb{E}[ ( \hat{\tau}^{\pi}_W - \tau^{\pi}_{SAPE} )^2 \mid X, T, Z \mid X, T ] \nonumber \\
&= \mathbb{E}[ \mathbb{E}[ \hat{\tau}^{\pi}_W - \tau^{\pi}_{SAPE} \mid X, T, Z]^2  \mid X, T ] + \mathbb{E}[ \mathbb{V}[ \hat{\tau}^{\pi}_W - \tau^{\pi}_{SAPE} \mid X, T, Z] \mid X, T] \nonumber \\
&= \mathbb{E}[ ( \frac{1}{n} \sum_{i,t} f_{it} \mu_t(Z_i) )^2 \mid X, T ] + \mathbb{E}[ \mathbb{V}[ \frac{1}{n} \sum_i W_i Y(T_i)  \mid X, T, Z] \mid X, T] \nonumber \\
&\leq \mathbb{E}[ ( \frac{1}{n} \sum_{i,t} f_{it} \mu_t(Z_i) )^2 \mid X, T ] + \frac{\sigma^2}{n^2} \|W\|_2^2 \nonumber
\end{align}
where $\sigma$ is the bound defined in \cref{asm:variance}.

Next observe that for any (possibly correlated) random variables $A_1,\ldots,A_n$ and numbers $p_1,\ldots,p_n \in \R n$ such that $\sum_ip_i=1$, we have $\mathbb{V}[\sum_i p_i A_i] \leq \max_i \mathbb{V}[A_i]$. Given this, we can simplify the first term above further, as follows:
\begin{align}
\mathbb{E}[ ( \frac{1}{n} \sum_{i,t} f_{it} \mu_t(Z_i) )^2 \mid X, T ] &= (\frac{1}{n} \sum_{i,t} f_{it} \mathbb{E}[\mu_t(Z_i) \mid X_i, T_i])^2 + \mathbb{V}[\frac{1}{n} \sum_{i,t} f_{it} \mu_t(Z_i) \mid X, T] \nonumber \\
&= (\frac{1}{n} \sum_{i,t} f_{it} \nu_t(X_i,T_i))^2 + \frac{1}{n^2} \sum_i \mathbb{V}[ \sum_t f_{it} \mu_t(Z_i) \mid X, T] \nonumber \\
&\leq (\frac{1}{n} \sum_{i,t} f_{it} \nu_t(X_i,T_i))^2 + \frac{1}{n^2} \sum_i \max_t f_{it}^2 \mathbb{V}[\mu_t(Z_i) \mid X, T] \nonumber \\
&\leq (\frac{1}{n} \sum_{i,t} f_{it} \nu_t(X_i,T_i))^2 + \frac{2\sigma^2}{n^2} \sum_i \max_t W_i^2 \delta_{T_i t} + \pi_t(X_i)^2 \nonumber \\
&\leq (\frac{1}{n} \sum_{i,t} f_{it} \nu_t(X_i,T_i))^2 + \frac{2\sigma^2}{n^2} \sum_i (W_i^2 + 1) \nonumber \\
&= (\frac{1}{n} \sum_{i,t} f_{it} \nu_t(X_i,T_i))^2 + \frac{2\sigma^2}{n^2} \|W\|_2^2 + \frac{2 \sigma^2}{n} \nonumber
\end{align}
where for the second inequality we used the fact that $(x+y)^2 \leq 2x^2 + 2y^2$. This gives us $CMSE(W, \mu) \leq J(W, \mu) + O_p(1/n)$, and combining this with the above gives $\mathbb{E}[(\hat{\tau}^{\pi}_W~-~\tau^{\pi})^2~\mid~X_{1:N},~T_{1:N}] \leq 2J(W, \mu) + O_p(1/n)$ as required.

\subsection{Proof of \cref{lem:gbound}}
\label{apx:gbound}

First, we will use the notation $f(z;x,t)$ for the conditional measure of $Z$ given $X=x$ and $T=t$, and observe that according to Bayes rule we have:
\begin{equation}
\frac{f(z;x,t'')}{f(z;x,t')} = \frac{e_{t''}(z)}{e_{t'}(z)} \frac{\eta_{t'}(x)}{\eta_{t''}(x)} \nonumber
\end{equation}

Define $\mathbb{E}_{xt}$ and $\mathbb{P}_{xt}$ as shorthand for expectation and probability given $X=x,T=t$ respectively. Then given the above, for any $M > 0$ we can bound
\begin{align*}
\nu_t(x,t'') &= \frac{\eta_{t''}(x)}{\eta_{t'}(x)} \mathbb{E}[\mu_t(Z) \mid X=x, T=t''] \\
&= \int_{\mathcal{Z}} f(z;x,t'') \mu_t(z) dz \\
&= \frac{\eta_{t''}(x)}{\eta_{t'}(x)} \int_{\mathcal{Z}} \frac{e_{t''}(z)}{e_{t'}(z)} \frac{\eta_{t'}(x)}{\eta_{t''}(x)} f(z;x,t') \mu_t(z) dz \\
&= \frac{\eta_{t'}(x)}{\eta_{t''}(x)} \int_{\mathcal{Z}} \frac{e_{t''}(z)}{e_{t'}(z)} f(z;x,t') \mu_t(z)  dz \\
&\leq \frac{\eta_{t'}(x)}{\eta_{t''}(x)} \left( M \mathbb{E}_{xt'}[\indicator{\frac{e_{t''}(z)}{e_{t'}(z)} \leq M} \mu_t(Z)] + \mathbb{E}_{xt'}[\indicator{\frac{e_{t''}(z)}{e_{t'}(z)} > M} \frac{e_{t''}(z)}{e_{t'}(z)} \mu_t(Z)] \right) \\
\end{align*}

Now we can use the fact that $\mu_t$ is $b$-bounded to bound the first term by
\begin{align*}
M \mathbb{E}_{xt'}[\indicator{\frac{e_{t''}(z)}{e_{t'}(z)} \leq M} \mu_t(Z)] &= M \nu_t(x,t') - M\mathbb{E}_{xt'}[\indicator{\frac{e_{t''}(z)}{e_{t'}(z)} > M} \mu_t(Z)] \\
&\leq M \nu_t(x,t') + Mb\mathbb{P}_{xt'}[\frac{e_{t''}(z)}{e_{t'}(z)} > M],
\end{align*}
and in addition applying Cauchy Schwartz we can bound the second term by
\begin{align*}
\mathbb{E}_{xt'}[\indicator{\frac{e_{t''}(z)}{e_{t'}(z)} > M} \frac{e_{t''}(z)}{e_{t'}(z)} \mu_t(Z)]  &\leq \sqrt{\mathbb{E}_{xt'}[\indicator{\frac{e_{t''}(z)}{e_{t'}(z)} > M} \mu_t(Z)^2] \mathbb{E}_{xt'}[\left(\frac{e_{t''}(z)}{e_{t'}(z)}\right)^2]} \\
&\leq \sqrt{b^2 \mathbb{P}_{xt'}[\frac{e_{t''}(z)}{e_{t'}(z)} > M] \mathbb{E}_{xt'}[\left(\frac{e_{t''}(z)}{e_{t'}(z)}\right)^2]} \\
&\leq b \sqrt{\mathbb{P}_{xt'}[\frac{1}{e_{t'}(z)} > M] \mathbb{E}_{xt'}[\left(\frac{1}{e_{t'}(z)}\right)^2]}.
\end{align*}

Now define $g(x,t) = \mathbb{E}_{xt}[\left(\frac{1}{e_t(z)}\right)^2]$. By \cref{asm:overlap} we know $g(x,t)$ is finite for every $x$ and $t$. Also by Markov's inequality we know that $\mathbb{P}_{xt}[\frac{1}{e_t(z)} > M] \leq \frac{g(x,t)}{M^2}$. Therefore putting all of the above together we can obtain
\begin{align}
\nu_t(x,t'') &\leq \frac{\eta_{t'}(x)}{\eta_{t''}(x)} \left(M \nu_t(x,t') + \frac{2bg(x,t')}{M} \right) \nonumber \\
&\leq \frac{\eta_{t'}(x)}{\eta_{t''}(x)} \left(M |\nu_t(x,t')| + \frac{2bg(x,t')}{M} \right) \nonumber 
\end{align}
This inequality is valid every $M$, so we can pick $M$ to make it as tight as possible. Choosing $M = \sqrt{\frac{2bg(x,t')}{\nu_t(x,t')}}$ gives us:
\begin{equation}
\nu_t(x,t'') \leq \frac{\eta_{t'}(x)}{\eta_{t''}(x)} \sqrt{8bg(x,t')|\nu_t(x,t')|} \nonumber 
\end{equation}

Finally note that, since by symmetry $\mathbb{E}[\mu(Z) \mid X,T] = -\mathbb{E}[-\mu(Z) \mid X,T]$, we can strengthen this inequality to the following
\begin{equation*}
|\nu_t(x,t'')| \leq \frac{\eta_{t'}(x)}{\eta_{t''}(x)} \sqrt{8bg(x,t')|\nu_t(x,t')|},
\end{equation*}
and noting that $g(x,t') = \mathbb{E}[e^{-2}_t(Z) \mid X=x, T=t']$ gives us our final result.





\subsection{Proof of \cref{lem:minimax}}
\label{apx:minimax}

First note that by \cref{asm:compact} $\mathcal{F}$ is compact. Also $J(W,\cdot)$ is continuous for every $W$, since by \cref{asm:ltwo} we know that the norm on each $\mathcal F_t$ dominates the norm on $\ltwo(\mathcal Z)$ and this continuity result would be trivial if $\mathcal F_t = \ltwo(\mathcal Z)$. This means that by the Extreme Value theorem we can replace the supremum over $\mu$ with a maximum over $\mu$ in the quantity we are bounding. Given this, we will proceed by bounding $\min_W \max_{\mu \in \mathcal{F}} B(W, \mu)$ using von Neumann's minimax theorem to swap the minimum and the maximum, and then use this to establish the overall bound for $J(W,\mu)$. 

Next we can observe that $B(W, \mu)$ is linear, and therefore both convex and concave, for each of $W$ and $\mu$. Next, by \cref{asm:compact} $\mathcal{F}$ is convex and compact, and following the same argument as above $B(W,\cdot)$ is continuous for every $W$. In addition, $B(W, \mu)$ is also clearly continuous in $W$ for fixed $\mu$, and the set $\{W : \|W\|_2 \leq M\}$ is obviously compact and convex for any constant $M$. Thus by von Neumann's minimax theorem we have the following for every finite $M$:
\begin{equation}
\label{eq:minimax}
\min_{\|W\|_2 \leq M} \max_{\mu \in \mathcal{F}} B(W, \mu) = \max_{\mu \in \mathcal{F}} \min_{\|W\|_2 \leq M} B(W, \mu)
\end{equation}

Given this, we can bound $\min_W \max_{\mu \in \mathcal{F}} \hat{J}(W, \mu)$ as follows, which is valid for any $M$:
\begin{align}
\min_{W} \max_{\mu \in \mathcal{F}} \hat{J}(W, \mu) &\leq \min_{W} \max_{\mu \in \mathcal{F}} B(W, \mu)^2 + \frac{1}{n^2} W^T \Sigma W \nonumber \\
&\leq \min_{W} \max_{\mu \in \mathcal{F}} B(W, \mu)^2 + \frac{\sigma^2}{n^2} \|W\|^2_2 \nonumber \\
&\leq \min_{\|W\|_2 \leq M} \max_{\mu \in \mathcal{F}} B(W, \mu)^2 + \frac{\sigma^2}{n^2}\|W\|^2_2 \nonumber \\
&\leq (\min_{\|W\|_2 \leq M} \max_{\mu \in \mathcal{F}} B(W, \mu))^2 + \frac{\sigma^2}{n^2}M^2 \nonumber \\
&= (\max_{\mu \in \mathcal{F}} \min_{\|W\|_2 \leq M} B(W, \mu))^2 + \frac{\sigma^2}{n^2}M^2 \nonumber \\
&= \max_{\mu \in \mathcal{F}} \min_{\|W\|_2 \leq M} B(W, \mu)^2 + \frac{\sigma^2}{n^2}M^2 \nonumber
\end{align}

In these inequalities we use the fact that $\min_W \max_{\mu} B(W, \mu)^2 = (\min_W \max_{\mu} B(W, \mu))^2$ and $\max_{\mu} \min_W B(W, \mu)^2 = (\max_{\mu} \min_W B(W, \mu))^2$ due to the symmetry of $\mu$ in $\mathcal{F}$ implied by \cref{asm:symmetry}.


\subsection{Proof of \cref{lem:jconvergence}}
\label{apx:jconvergence}

Let $\prod$ denote Cartesian product. First we note that without loss of generality we can prove this lemma in the case that $\mathcal F = \prod_t \mathcal F_t$, since in general $F \subseteq \prod_t \mathcal F_t$ so $\sup_{\mu \in \mathcal F} \inf_W J(W,\mu) \leq \sup_{\mu \in \prod_t \mathcal F_t} \inf_W J(W,\mu)$, and it is easy to verify that all of our assumptions would still hold on the larger set $\prod_t \mathcal F_t$. 

Now define the set $\mathcal{H}^0 = \{\mu \in \prod_t \ltwo(\mathcal Z) : \mathbb{E}[\nu_T(X,T)^2] = 0\}$. Each coordinate $\mathcal H_t^0$ of $\mathcal H^0$ is a subspace of $\ltwo(\mathcal Z)$, so we can also define its orthogonal complement $\mathcal{H}_t^+$. Also, we have separability since from \cref{sec:setting} we know that $\mathcal Z \subseteq \mathbb{R}^q$, so any function $f \in \ltwo(\mathcal Z)$ can be uniquely represented as $f = f^0 + f^+$ where $f^0 \in \mathcal{H}_t^0$ and $f^+ \in \mathcal{H}_t^+$. This means that for each $\mu_t \in \mathcal F_t$, we can similarly uniquely represent $\mu_t = \mu_t^0 + \mu_t^+$, and we can easily extend this to a unique representation of the vector $\mu = \mu^0 + \mu^+$. Now in the case that $\mathbb{E}[\nu_T(X,T)^2] = 0$ we have $\nu_{T_i}(X_i,T_i) = 0$ almost surely for all $i$, and it follows from \cref{lem:gbound} that $\nu_{T_i}(X_i,t) = 0$ almost surely also for all $i$ and $t$. Therefore any component of $\mu$ in $\mathcal{H}^0$ has no effect on the function $J(W,\mu)$ which we are bounding, so without loss of generality we can restrict our attention to the following space:
\begin{equation*}
\mathcal{F}^+ = \prod_t (\mathcal{H}_t^+ \cap \mathcal F_t).
\end{equation*}

By construction the only function in $\mathcal{F}^+$ such that $\mathbb{E}[\nu_T(X,T)^2] = 0\}$ is the zero function, which we can also ignore in our bounds below, since when $\mu = 0$ we can easily obtain $J(W,\mu) = 0$ by choosing $W = 0$. Furthermore, by \cref{asm:ltwo} we know that for each $t$ the $\mathcal F_t$ norm dominates the $\ltwo(\mathcal Z)$ norm, so it must be the case that that each space $\mathcal{F}_t^+$ is closed, since $\mathcal{H}_t^+$ is a closed subspace of $\ltwo(\mathcal Z)$ due to it being an orthogonal complement. Thus it follows easily from \cref{asm:norm} that $\mathcal{F}^+$ is closed, given that its norm is an $\mathbb{R}^m$ norm on top of the corresponding $\mathcal{F}^+_t$ norms and $m$ is finite.

Now, based on \cref{lem:minimax}, it is sufficient to pick weights in response to $\mu$ that control for a single mean outcome function. Instead of actually constructing a particular set of weights, we take the approach of viewing this as a convex optimization problem. Specifically, given $\mu$, we calculate the minimum euclidean norm of all weights that set the bias term $B(W,\mu)$ to zero exactly. This can be formulated as the following convex optimization program
\begin{align*}
&\min_W \sum_i W_i^2 \\
&\text{s.t.} \sum_i W_i \nu_{T_i}(X_i, T_i) = \sum_{i,t} \pi_t(X_i) \nu_t(X_i, T_i).
\end{align*}

Given the program only has linear constraints with equality, it satisfies Slater's condition, and therefore satisfies strong duality, which we will use to find the optimal value of this program. First we calculate the Lagrangian as
\begin{equation*}
\mathcal L_n(W, \lambda) = \sum_i W_i^2 + \lambda \left( \sum_i W_i \nu_{T_i}(X_i, T_i) - \sum_{i,t} \pi_t(X_i) \nu_t(X_i, T_i) \right).
\end{equation*}

It can easily be verified by taking derivatives that for any $\lambda \in \mathbb{R}$ this function is minimized by setting $W_i = -\frac{\lambda}{2} \nu_{T_i}(X_i,T_i)$, and plugging this value in gives the dual formulation of the program as
\begin{align*}
&\max_{\lambda} -\frac{\lambda^2}{4} \sum_i \nu_{T_i}(X_i,T_i)^2 - \lambda \sum_{i,t} \pi_t(X_i) \nu_t(X_i,T_i),
\end{align*}
which is unconstrained. Again by taking derivatives we can maximize this function, and we find the maximum value is given by
\begin{equation*}
\lambda = -2\frac{\sum_i \pi_t(X_i) \nu_t(X_i, T_i)}{\sum_i \nu_{T_i}(X_i, T_i)^2},
\end{equation*}
and finally plugging this value into the dual objective function we see that the euclidean norm of the weights $W^*$ solving the convex program above is given by
\begin{equation*}
\|W^*\|_2^2 = \frac{\left(\sum_i \pi_t(X_i) \nu_t(X_i, T_i)\right)^2}{\sum_i \nu_{T_i}(X_i, T_i)^2}.
\end{equation*}

Now define $\mathbb{E}_n$ as the mean with respect to the empirical distribution of the logged data. Then this objective value can be reformulated as
\begin{equation*}
\|W^*\|_2^2 = \frac{n\mathbb{E}_n[\sum_t \nu_t(X,T)]^2}{\mathbb{E}_n[\nu_{T}(X,T)^2]}.
\end{equation*}

Therefore choosing $M = \|W^*\|$, combining this result with \cref{lem:minimax} gives us
\begin{align*}
\min_W \sup_{\mu \in \mathcal{F}^+} J(W, \mu) &\leq \sup_{\mu \in \mathcal{F}^+} \frac{1}{n} \left( \frac{\sigma^2\mathbb{E}_n[\sum_t \nu_t(X,T)]^2}{\mathbb{E}_n[\nu_{T}(X,T)^2]} \right) \\
&= \sup_{\mu \in \mathcal{F}^+} \frac{1}{n} \left( \frac{\sigma^2\mathbb{E}_n[\sum_t \nu_t(X,T)]^2}{\mathbb{E}[\nu_{T}(X,T)^2] + (\mathbb{E}_n[\nu_{T}(X,T)^2] - \mathbb{E}[\nu_{T}(X,T)^2])} \right).
\end{align*}

Given this we will proceed by arguing that we can bound the denominator away from zero. We can note that $\mu$ appears in both the numerator and denominator on the same scale, so without loss of generality we can further restrict our attention to $\mu$ with fixed norm. By \cref{asm:nondegen} we know that we can rescale every $\mu \in \mathcal F$ to have norm $\gamma$ for some $\gamma > 0$. Given this we will restrict ourselves to the set $\mathcal{F}^+_{\gamma} = \{\mu \in \mathcal{F}^+ : \|\mu\| = \gamma\}$. Since $\mathcal{F}^+_{\gamma}$ is the intersection of two closed sets it must be closed. Furthermore by \cref{asm:compact} it is also compact, so it satisfies the conditions for the extreme value theorem. By construction $\mathbb{E}[\nu_T(X,T)^2] > 0$ for every $\mu \in \mathcal{F}^+_{\gamma}$, so putting the above together we have $\inf_{\mu \in \mathcal{F}^+_{\gamma}} \mathbb{E}[\nu_T(X,T)^2] > 0$. We will define this value to be $\alpha$. 

Now the numerator in the above bound is clearly bounded above by some $\beta > 0$ uniformly over $\mu \in \mathcal F$, since by \cref{asm:bounded} we know that every $\mu_t \in \mathcal F_t$ is uniformly bounded by some global constant, and therefore all $\nu$ terms are bounded by some constant $b$. Given this all that remains to be shown is that $\sup_{\mu \in \mathcal F} |\mathbb{E}_n[\nu_{T}(X,T)^2] - \mathbb{E}[\nu_{T}(X,T)^2]|$ converges in probability to zero. In order to show this we will define the following terms:
\begin{align*}
D_n &= \mathbb{E}_n[\nu_T(X,T)^2] \\
E_n &= \sup_{\mu \in \mathcal{F}} |D_n - \mathbb{E}[D_n]|
\end{align*}

We need to show that $E_n$ converges uniformly to zero. Define $D_n'$ as an arbitrary recalculation of $D_n$ replacing $(X_{1:n}, T_{1:n})$ with $(X'_{1:n}, T'_{1:n})$, which differ from the originals at most in a single coordinate $i$, and define $E_n' = \sup_{\mu \in \mathcal{F}} |D_n' - \mathbb{E}[D_n']|$. Furthermore as argued above all $\nu$ terms are bounded above by some constant $b$, so each $\nu_{T_i}(X_i,T_i)^2$ is bounded by $b^2$. Given this we can obtain
\begin{align*}
|E_n - E_n'| &= |\sup_{\mu \in \mathcal{F}} |D_n - \mathbb{E}[D_n]| - \sup_{\mu \in \mathcal{F}} |D_n' - \mathbb{E}[D_n']| | \\
&\leq \sup_{\mu \in \mathcal{F}} |(D_n - \mathbb{E}[D_n]) - (D_n' - \mathbb{E}[D_n'])| \\
&= \sup_{\mu \in \mathcal{F}} |D_n - D_n'| \\
&= \frac{1}{n} \sup_{\mu \in \mathcal{F}} |\nu_{T_i}(X_i,T_i)^2 - \nu_{T_i'}(X_i',T_i')^2| \\
&\leq \frac{2b^2}{n}
\end{align*}
Given this we can apply McDiarmid's inequality to obtain the following bound:
\begin{equation*}
\label{eq:mcdiarmid}
P(|E_n - \mathbb{E}[E_n]| \leq \epsilon) \leq 2\exp \left( -\frac{n\epsilon^2}{2b^4} \right) 
\end{equation*}

This implies that $E_n - \mathbb{E}[E_n] = o_p(1)$. Next we show that $\mathbb{E}[E_n] = o_p(1)$ also. We do this using a symmetrization argument as follows, where $D'_n$ is defined identically to $D_n$ using iid shadow variables $(X'_i,T'_i)$ in place of $(X_i,T_i)$ for each $i$, and $\epsilon_i$ are iid Rademacher random variables:
\begin{align*}
\mathbb{E}[E_n] &= \mathbb{E}\left[ \sup_{\mu \in \mathcal{F}} \left| D_n - \mathbb{E}[D_n] \right| \right] \\
&= \mathbb{E} \left[ \sup_{\mu \in \mathcal{F}} \left| \frac{1}{n} \sum_i \nu_{T_i}(X_i,T_i) - \mathbb{E}[\nu_{T_i'}(X_i',T_i')] \right| \right] \\
&\leq 2 \mathbb{E} \left[ \sup_{\mu \in \mathcal{F}} \left| \frac{1}{n} \sum_i \epsilon_i \nu_{T_i}(X_i,T_i)  \right| \right]  \\
&\leq 2 \sum_t \mathbb{E} \left[ \sup_{\mu \in \mathcal{F}} \left| \frac{1}{n} \sum_i \epsilon_i \delta_{T_i t} \nu_{t}(X_i,T_i)  \right| \right] \\
&\leq 4 \sum_t \mathbb{E} \left[ \sup_{\mu \in \mathcal{F}} \left| \frac{1}{n} \sum_i \epsilon_i \nu_{t}(X_i,T_i)  \right| \right] \\
&\leq 4 \sum_t \mathbb{E} \left[ \sup_{\mu \in \mathcal{F}} \left| \frac{1}{n} \sum_i \epsilon_i \mu_t(Z_i)  \right| \right] \\
&\leq 4 \sum_t \mathcal R_n(\mathcal F_t) \\
\end{align*}
where in the third inequality we appeal to the Rademacher comparison lemma \citep[Thm.~4.12]{ledoux2013probability}. Thus since from \cref{asm:complexity} we know that the Rademacher complexity of each set $\mathcal R_n(\mathcal F_n)$ vanishes, it follows that $\mathbb{E}[E_n] = o_p(1)$. Putting everything from above together we get
\begin{equation*}
\min_W \sup_{\mu \in \mathcal{F}} J(W, \mu) \leq \frac{1}{n} \frac{\beta}{\alpha + o_p(1)},
\end{equation*}
so we have $\min_W \sup_{\mu \in \mathcal{F}} J(W, \mu) \leq O_p(1/n)$ as required.

\subsection{Proof of \cref{thm:fk}}
\label{apx:fk}

First, \cref{asm:norm} follows trivially from the definition of $\fk$. Next, \cref{asm:symmetry} and $\cref{asm:nondegen}$ follow from the fact that $\fk$ consists of all functions in $\op{span}(\mathcal{F}_t)$ with norm at most 1, as does the fact that it is a closed space. Given that $K$ is a Mercer kernel, balls in the corresponding RKHS have finite covering number \citep{zhou2002covering}, and it follows easily from this that $\fk$ has finite covering numbers, as its covering number must be bounded above by the sum of the covering numbers of the spaces $\fkt$. So $\fk$ is closed and totally bounded with respect to its norm, and therefore compact, which gives us \cref{asm:compact}. Clearly each $\mathcal F_t$ is contained in $\ltwo(\mathcal Z)$ since RKHS spaces are square integrable, and the fact that the $K$ norm dominates the $\ltwo$ follows from Mercer's Theorem, which implies that $\|f\|_K^2 = \sum_{i=1}^{\infty} f_i^2 / \sigma_i$, where $f_i$ is the $i$th eigenvalue of $f$ for some orthonormal basis of $\ltwo(\mathcal Z)$, and $\sigma_i \geq 0$ converges to zero. This gives us \cref{asm:ltwo}. Next, by construction each $\fkt$ consists of all functions in the RKHS up to norm 1. Therefore assuming the kernel $K$ is bounded, it is trivial to verify that function application must be globally bounded, since for any function $f \in \fkt$ we have $f(x) \leq <f, K_x> \leq \|f\| \sqrt{K(x,x)} \leq \sqrt{K(x,x)}$, which gives us \cref{asm:bounded}. Finally, given this characterization of $\fkt$ as the 1-ball of the RKHS, it has vanishing Rademacher complexity \citep[Thm.~2.1]{mendelson2003performance}, so we have \cref{asm:complexity}.


\subsection{Proof of \cref{thm:continuous-algorithm}}
\label{apx:continuous-algorithm}

First we will find a closed form expression for $\sup_{\mu \in \fk} (\frac{1}{n} \sum_{i,t} f_{it} \nu_{t}(X_i,T_i))^2$. In this derivation we will use the shorthand $\varphi_{ti}$ for the conditional density of $\mu_t$ given $X_i$ and $T_i$, and $T_K$ for the kernel intergral operator defined according to $T_K f = \int_{\mathcal Z} K(\cdot,z) f(z) dz$. In this derivation we will make use of the fact that $\ltwoprod{f}{g} = \rkhsprod{f}{T_K g}$ for any square integrable $f$ and $g$. Given all this we can obtain:
\begin{align*}
\sup_{\mu \in \fk} \left( \frac{1}{n} \sum_{i,t} f_{it} \nu_{t}(X_i,T_i) \right)^2 &= \sum_t \sup_{\mu_t \in \fkt} \left( \frac{1}{n} \sum_i f_{it} \ltwoprod{\mu_t}{\varphi_i} \right)^2 \\
&= \sum_t \sup_{\mu_t \in \fkt} \ltwoprod{\mu_t}{\frac{1}{n} \sum_i f_{it} \varphi_i}^2 \\
&= \sum_t \sup_{\mu_t \in \fkt} \rkhsprod{\mu_t}{T_K \frac{1}{n} \sum_i f_{it} \varphi_i}^2 \\
&= \sum_t \frac{\rkhsprod{T_K \frac{1}{n} \sum_i f_{it} \varphi_i}{T_K \frac{1}{n} \sum_i f_{it} \varphi_i}^2}{\|T_K \frac{1}{n} \sum_i f_{it} \varphi_i\|_K} \\
&= \sum_t \rkhsprod{T_K \frac{1}{n} \sum_i f_{it} \varphi_i}{T_K \frac{1}{n} \sum_i f_{it} \varphi_i} \\
&= \sum_t \ltwoprod{\frac{1}{n} \sum_i f_{it} \varphi_i}{T_K \frac{1}{n} \sum_i f_{it} \varphi_i} \\
&= \frac{1}{n^2} \sum_{i,j,t} f_{it} f_{jt} \ltwoprod{\varphi_i}{T_K \varphi_j} \\
&= \frac{1}{n^2} \sum_{i,j,t} f_{it} f_{jt} \int_{\mathcal Z} \varphi_i(z) (\int_{\mathcal Z'} K(z,z') \varphi_j(z') dz') dz \\
&= \frac{1}{n^2} \sum_{i,j,t} f_{it} f_{jt} \int_{\mathcal Z} \int_{\mathcal Z'} \varphi_i(z) \varphi_j(z') K(z,z') dz' dz \\
&= \frac{1}{n^2} \sum_{i,j,t} f_{it} f_{jt} \mathbb{E}[K(Z_i,Z_j')] 
\end{align*}

Given this, and recalling that $f_{it} = W_i \delta_{T_i t} - \pi_t(X_i)$ we can derive a closed form for our minimization objective, as follows:
\begin{align*}
\sup_{\mu \in \fk} J(W, \mu) &= \frac{1}{n^2} \sum_{i,j,t}  f_{it} f_{jt} Q_{ij} + \frac{1}{n^2} \sum_{i,j} W_i W_j \Gamma_{ij} \\
&= \frac{1}{n^2} \sum_{i,j,t}  Q_{ij} (W_i W_j \delta_{T_i t} \delta_{T_j t} - 2 W_j \delta_{T_j t} \pi_t(X_i) + \pi_t(X_i) \pi_t(X_j)) \\
&\qquad\qquad + \frac{1}{n^2} \sum_{i,j} W_i W_j \Gamma_{ij} \\
&= \frac{1}{n^2} \sum_{i,j} W_i W_j (Q_{ij} \delta_{T_i T_j} + \Gamma_{ij}) - \frac{2}{n^2} \sum_j W_j (\sum_i Q_{ij} \pi_{T_j}(X_i)) \\
&\qquad\qquad + \frac{1}{n^2} \sum_{i,j,t} Q_{ij} \pi_t(X_i) \pi_t(X_j)
\end{align*}

Finally we can conclude by noting that this corresponds to the quadratic program formulation given in the question with $c = \frac{1}{n^2} \sum_{i,j,t} Q_{ij} \pi_t(X_i) \pi_t(X_j)$.





\section{Additional Experimentation Details}

\subsection{Experiment Scenario}
\label{apx:exp-scenario}

All our experiments were conducted using the setup described in \cref{sec:exp-setup}. We used the following parameter values for our data-generating distribution:
\begin{align*}
\alpha &= [1.0, -2.0, -1.0, 2.0, 4.0, 0.0, -2.0, -1.0, -3.0, 1.0] \\
\alpha_0 &= 0 \\
\sigma^2_X &= 4.0 \\
\beta &= [0.5, -0.5] \\
\beta_0 &= 0 \\
\zeta(0) &= 1.0 \\
\zeta(0)_0 &= 0 \\
\zeta(1) &= -0.5 \\
\zeta(1)_0 &= 0 \\
\sigma^2_Y &= 0.01 
\end{align*}

In addiiton, the policy $\pi$ we are evaluating takes the form as described in \cref{sec:exp-setup}, and we used the following parameter values for this policy:
\begin{align*}
\psi_0 &= [-0.1, 0.2, 0.2, -0.1, -0.1, -0.1, 0.1, 0.1, 0.1, -0.1] \\
\psi_1 &= - \psi_0 
\end{align*}

\subsection{Method Implementation Details}
\label{apx:exp-benchmark}

In all methods where we sampled from the posterior $\varphi(\cdot;x,t)$, this sampling was done using STAN \citep{carpenter2017stan}, solving QPs and LCQPs was done using the Python package \texttt{quadprog},\footnote{https://pypi.org/project/quadprog/} all stochastic gradient descent (SGD) learning was performed using the Adam \citep{kingma2014adam} optimizer with a learning rate of 0.001.

\paragraph{\textbf{OptZ}} We ran \cref{algo:continuous} with $B = 50$.

\paragraph{\textbf{IPS}} Since the propensity scores $\eta_t(x)$ are not not tractable to compute analytically, we trained a neural network $\hat \eta$ to estimate this function. This was done by sampling batches of $(Z,X)$ pairs from the data model, and training the network using SGD to predict the vector of probabilities $P_T = \beta^T Z + \beta_0$ from $X$, using cross-entropy loss. We used a neural network with two hidden layers of size $200$ for $\hat \eta$, and trained for $2000$ iterations with a batch size of 32. We found in practice this training was very stable and gave accurate results. 

\paragraph{\textbf{DirX}} For each $t$ we trained a neural network $\hat \rho_t$ to predict $\nu_t(x,t)$ by taking the set of $(X,T,Y)$ triplets in our training data where $T=t$, and training the network using SGD to predict $Y$ from $X$ using MSE loss. Based on pilot experiments we used a network architecture with a single hidden layer of size $100$, and trained using a batch size of 128. We used $80\%$ of our data for training, and used the remaining $20\%$ for the purpose of early stopping. We trained for a maximum of $500$ epochs, or until we made no progress on development data for $20$ epochs. 

\paragraph{\textbf{DirZ}} For each $t$ we trained a neural network $\hat \mu_t$ to predict $\mu_t$. This was done by taking the set of $(X,T,Y)$ triplets in our training data, and for each sampling $200$ $Z$ values from the posterior using our identified model given $X$ and $T$. This gives us a set of $(Z,T,Y)$ triplets $200$ times as large as our original training set. We then trained each $\hat \mu_t$ network by taking the set of these triplets where $T=t$, and optimized the network using SGD on this data predicting $Y$ from $Z$. We used the same settings for this optimization as with the \emph{direct-naive} method, except we allowed up to $1000$ epochs. Note that for both training and inference we limited ourselves to sampling $200$ $Z$ values per data point due to computational limitations.

\section{Additional Experiment Results}
\label{apx:exp-results}

We present here our additional experiment results. In these results \textbf{SimplexOptZ} refers to our method using the simplex constraints discussed in \cref{sec:qpalgo}.

\begin{table}
\begin{center}
\footnotesize
\begin{tabular}{ccccc}
\hline
n & $\textbf{OptZ}_{0.001}$ & $\textbf{OptZ}_{0.2}$ & $\textbf{OptZ}_{1.0}$ & $\textbf{OptZ}_{5.0}$\\
\hline
200 & $.39\pm.07$ & $.24\pm.02$ & $.36\pm.02$ & $.81\pm.02$\\
500 & $.19\pm.02$ & $.18\pm.02$ & $.23\pm.02$ & $.49\pm.02$\\
1000 & $.11\pm.01$ & $.11\pm.01$ & $.13\pm.01$ & $.27\pm.01$\\
2000 & $.08\pm.01$ & $.08\pm.01$ & $.09\pm.01$ & $.17\pm.01$\\
\hline
\end{tabular}
\end{center}
\caption{Convergence of RMSE for weighted estimator using our weights, with \textbf{step} link}
\label{tab:step-rmse}

\begin{center}
\footnotesize
\begin{tabular}{ccccc}
\hline
n & $\textbf{DirX:OptZ}_{0.001}$ & $\textbf{DirX:OptZ}_{0.2}$ & $\textbf{DirX:OptZ}_{1.0}$ & $\textbf{DirX:OptZ}_{5.0}$\\
\hline
200 & $.57\pm.06$ & $.42\pm.03$ & $.39\pm.03$ & $.43\pm.03$\\
500 & $.55\pm.02$ & $.46\pm.02$ & $.39\pm.02$ & $.37\pm.02$\\
1000 & $.49\pm.02$ & $.45\pm.01$ & $.39\pm.01$ & $.32\pm.01$\\
2000 & $.48\pm.01$ & $.47\pm.01$ & $.42\pm.01$ & $.34\pm.01$\\
\hline
\end{tabular}
\end{center}
\caption{Convergence of RMSE for doubly robust estimator using our weights and $\textbf{DirX}$, with \textbf{step} link}
\label{tab:step-rmse-drx}

\begin{center}
\footnotesize
\begin{tabular}{ccccc}
\hline
n & $\textbf{DirZ:OptZ}_{0.001}$ & $\textbf{DirZ:OptZ}_{0.2}$ & $\textbf{DirZ:OptZ}_{1.0}$ & $\textbf{DirZ:OptZ}_{5.0}$\\
\hline
200 & $.41\pm.07$ & $.29\pm.02$ & $.50\pm.02$ & $1.1\pm.03$\\
500 & $.20\pm.02$ & $.21\pm.02$ & $.31\pm.02$ & $.70\pm.02$\\
1000 & $.11\pm.01$ & $.13\pm.01$ & $.18\pm.01$ & $.42\pm.01$\\
2000 & $.08\pm.01$ & $.09\pm.01$ & $.13\pm.01$ & $.26\pm.01$\\
\hline
\end{tabular}
\end{center}
\caption{Convergence of RMSE for doubly robust estimator using our weights and $\textbf{DirZ}$, with \textbf{step} link}
\label{tab:step-rmse-drz}

\begin{center}
\footnotesize
\begin{tabular}{ccccc}
\hline
n & $\textbf{SimplexOptZ}_{0.001}$ & $\textbf{SimplexOptZ}_{0.2}$ & $\textbf{SimplexOptZ}_{1.0}$ & $\textbf{SimplexOptZ}_{5.0}$\\
\hline
200 & $.30\pm.02$ & $.25\pm.02$ & $.38\pm.02$ & $.91\pm.02$\\
500 & $.18\pm.02$ & $.19\pm.02$ & $.24\pm.02$ & $.54\pm.02$\\
1000 & $.12\pm.01$ & $.11\pm.01$ & $.13\pm.01$ & $.29\pm.01$\\
2000 & $.07\pm.01$ & $.08\pm.01$ & $.10\pm.01$ & $.18\pm.01$\\
\hline
\end{tabular}
\end{center}
\caption{Convergence of RMSE for weighted estimator using our weights and constraining $W \in n \Delta^n$, with \textbf{step} link}
\label{tab:step-rmse-simplex}

\begin{center}
\footnotesize
\begin{tabular}{cccccccc}
\hline
n & $\textbf{IPS}$ & $\textbf{OptX}_{0.001}$ & $\textbf{OptX}_{0.2}$ & $\textbf{OptX}_{1.0}$ & $\textbf{OptX}_{5.0}$ & $\textbf{DirX}$ & $\textbf{DirZ}$\\
\hline
200 & $.47\pm.03$ & $2.0\pm.03$ & $2.1\pm.03$ & $2.3\pm.02$ & $2.5\pm.02$ & $.52\pm.02$ & $2.6\pm.02$\\
500 & $.48\pm.03$ & $2.0\pm.02$ & $2.1\pm.02$ & $2.3\pm.02$ & $2.6\pm.02$ & $.48\pm.02$ & $2.6\pm.01$\\
1000 & $.39\pm.02$ & $2.0\pm.01$ & $2.1\pm.01$ & $2.3\pm.01$ & $2.5\pm.01$ & $.48\pm.02$ & $2.6\pm.01$\\
2000 & $.40\pm.01$ & $2.0\pm.01$ & $2.1\pm.01$ & $2.3\pm.01$ & $2.5\pm.01$ & $.45\pm.02$ & $2.6\pm.01$\\
\hline
\end{tabular}
\end{center}
\caption{Convergence of RMSE for benchmark methods, with \textbf{step} link}
\label{tab:step-rmse-bench}

\end{table}

\begin{table}
\begin{center}
\footnotesize
\begin{tabular}{ccccc}
\hline
n & $\textbf{OptZ}_{0.001}$ & $\textbf{OptZ}_{0.2}$ & $\textbf{OptZ}_{1.0}$ & $\textbf{OptZ}_{5.0}$\\
\hline
200 & $.03\pm.39$ & $.11\pm.21$ & $.29\pm.21$ & $.78\pm.18$\\
500 & $.09\pm.17$ & $.10\pm.15$ & $.17\pm.16$ & $.47\pm.15$\\
1000 & $.02\pm.11$ & $.05\pm.09$ & $.08\pm.09$ & $.25\pm.09$\\
2000 & $.03\pm.07$ & $.05\pm.06$ & $.07\pm.07$ & $.16\pm.07$\\
\hline
\end{tabular}
\end{center}
\caption{Convergence of bias for weighted estimator using our weights, with \textbf{step} link}
\label{tab:step-bias}

\begin{center}
\footnotesize
\begin{tabular}{ccccc}
\hline
n & $\textbf{DirX:OptZ}_{0.001}$ & $\textbf{DirX:OptZ}_{0.2}$ & $\textbf{DirX:OptZ}_{1.0}$ & $\textbf{DirX:OptZ}_{5.0}$\\
\hline
200 & $.43\pm.38$ & $.35\pm.24$ & $.31\pm.24$ & $.37\pm.22$\\
500 & $.51\pm.19$ & $.42\pm.18$ & $.35\pm.18$ & $.33\pm.17$\\
1000 & $.47\pm.13$ & $.44\pm.11$ & $.37\pm.10$ & $.30\pm.11$\\
2000 & $.47\pm.09$ & $.46\pm.08$ & $.41\pm.08$ & $.33\pm.08$\\
\hline
\end{tabular}
\end{center}
\caption{Convergence of bias for doubly robust estimator using our weights and $\textbf{DirX}$, with \textbf{step} link}
\label{tab:step-bias-drx}

\begin{center}
\footnotesize
\begin{tabular}{ccccc}
\hline
n & $\textbf{DirZ:OptZ}_{0.001}$ & $\textbf{DirZ:OptZ}_{0.2}$ & $\textbf{DirZ:OptZ}_{1.0}$ & $\textbf{DirZ:OptZ}_{5.0}$\\
\hline
200 & $.05\pm.40$ & $.19\pm.22$ & $.45\pm.22$ & $1.1\pm.21$\\
500 & $.10\pm.18$ & $.14\pm.16$ & $.26\pm.16$ & $.68\pm.17$\\
1000 & $.04\pm.11$ & $.09\pm.10$ & $.15\pm.10$ & $.41\pm.10$\\
2000 & $.03\pm.07$ & $.06\pm.07$ & $.10\pm.07$ & $.25\pm.07$\\
\hline
\end{tabular}
\end{center}
\caption{Convergence of bias for doubly robust estimator using our weights and $\textbf{DirZ}$, with \textbf{step} link}
\label{tab:step-bias-drz}

\begin{center}
\footnotesize
\begin{tabular}{ccccc}
\hline
n & $\textbf{SimplexOptZ}_{0.001}$ & $\textbf{SimplexOptZ}_{0.2}$ & $\textbf{SimplexOptZ}_{1.0}$ & $\textbf{SimplexOptZ}_{5.0}$\\
\hline
200 & $.04\pm.30$ & $.12\pm.21$ & $.31\pm.21$ & $.89\pm.20$\\
500 & $.08\pm.15$ & $.10\pm.15$ & $.18\pm.16$ & $.51\pm.16$\\
1000 & $.01\pm.12$ & $.06\pm.09$ & $.09\pm.09$ & $.27\pm.10$\\
2000 & $.03\pm.07$ & $.05\pm.06$ & $.07\pm.07$ & $.17\pm.07$\\
\hline
\end{tabular}
\end{center}
\caption{Convergence of bias for weighted estimator using our weights and constraining $W \in n \Delta^n$, with \textbf{step} link}
\label{tab:step-bias-simplex}

\begin{center}
\footnotesize
\begin{tabular}{cccccccc}
\hline
n & $\textbf{IPS}$ & $\textbf{OptX}_{0.001}$ & $\textbf{OptX}_{0.2}$ & $\textbf{OptX}_{1.0}$ & $\textbf{OptX}_{5.0}$ & $\textbf{DirX}$ & $\textbf{DirZ}$\\
\hline
200 & $.40\pm.25$ & $1.9\pm.21$ & $2.1\pm.20$ & $2.3\pm.19$ & $2.5\pm.18$ & $.49\pm.18$ & $2.6\pm.14$\\
500 & $.43\pm.21$ & $2.0\pm.16$ & $2.1\pm.15$ & $2.3\pm.14$ & $2.6\pm.13$ & $.45\pm.16$ & $2.6\pm.12$\\
1000 & $.37\pm.12$ & $2.0\pm.10$ & $2.1\pm.09$ & $2.3\pm.09$ & $2.5\pm.08$ & $.46\pm.15$ & $2.6\pm.11$\\
2000 & $.39\pm.10$ & $2.0\pm.08$ & $2.1\pm.07$ & $2.3\pm.07$ & $2.5\pm.07$ & $.42\pm.17$ & $2.6\pm.11$\\
\hline
\end{tabular}
\end{center}
\caption{Convergence of bias for benchmark methods, with \textbf{step} link}
\label{tab:step-bias-bench}

\end{table}

\begin{table}
\begin{center}
\footnotesize
\begin{tabular}{ccccc}
\hline
n & $\textbf{OptZ}_{0.001}$ & $\textbf{OptZ}_{0.2}$ & $\textbf{OptZ}_{1.0}$ & $\textbf{OptZ}_{5.0}$\\
\hline
200 & $.07\pm.01$ & $.04\pm.00$ & $.04\pm.00$ & $.07\pm.00$\\
500 & $.04\pm.00$ & $.03\pm.00$ & $.03\pm.00$ & $.04\pm.00$\\
1000 & $.02\pm.00$ & $.02\pm.00$ & $.02\pm.00$ & $.02\pm.00$\\
2000 & $.01\pm.00$ & $.01\pm.00$ & $.01\pm.00$ & $.01\pm.00$\\
\hline
\end{tabular}
\end{center}
\caption{Convergence of RMSE for weighted estimator using our weights, with \textbf{exp} link}
\label{tab:exp-rmse}

\begin{center}
\footnotesize
\begin{tabular}{ccccc}
\hline
n & $\textbf{DirX:OptZ}_{0.001}$ & $\textbf{DirX:OptZ}_{0.2}$ & $\textbf{DirX:OptZ}_{1.0}$ & $\textbf{DirX:OptZ}_{5.0}$\\
\hline
200 & $.13\pm.01$ & $.10\pm.01$ & $.12\pm.01$ & $.11\pm.01$\\
500 & $.10\pm.01$ & $.09\pm.01$ & $.10\pm.01$ & $.12\pm.01$\\
1000 & $.08\pm.00$ & $.08\pm.00$ & $.08\pm.00$ & $.10\pm.00$\\
2000 & $.07\pm.00$ & $.07\pm.00$ & $.08\pm.00$ & $.09\pm.00$\\
\hline
\end{tabular}
\end{center}
\caption{Convergence of RMSE for doubly robust estimator using our weights and $\textbf{DirX}$, with \textbf{exp} link}
\label{tab:exp-rmse-drx}

\begin{center}
\footnotesize
\begin{tabular}{ccccc}
\hline
n & $\textbf{DirZ:OptZ}_{0.001}$ & $\textbf{DirZ:OptZ}_{0.2}$ & $\textbf{DirZ:OptZ}_{1.0}$ & $\textbf{DirZ:OptZ}_{5.0}$\\
\hline
200 & $.15\pm.02$ & $.15\pm.01$ & $.25\pm.01$ & $.44\pm.01$\\
500 & $.10\pm.01$ & $.11\pm.01$ & $.18\pm.01$ & $.32\pm.01$\\
1000 & $.07\pm.01$ & $.08\pm.01$ & $.12\pm.01$ & $.23\pm.01$\\
2000 & $.04\pm.00$ & $.05\pm.00$ & $.08\pm.00$ & $.16\pm.00$\\
\hline
\end{tabular}
\end{center}
\caption{Convergence of RMSE for doubly robust estimator using our weights and $\textbf{DirZ}$, with \textbf{exp} link}
\label{tab:exp-rmse-drz}

\begin{center}
\footnotesize
\begin{tabular}{ccccc}
\hline
n & $\textbf{SimplexOptZ}_{0.001}$ & $\textbf{SimplexOptZ}_{0.2}$ & $\textbf{SimplexOptZ}_{1.0}$ & $\textbf{SimplexOptZ}_{5.0}$\\
\hline
200 & $.05\pm.00$ & $.11\pm.01$ & $.22\pm.01$ & $.39\pm.02$\\
500 & $.04\pm.00$ & $.08\pm.01$ & $.15\pm.01$ & $.28\pm.01$\\
1000 & $.02\pm.00$ & $.05\pm.00$ & $.09\pm.00$ & $.19\pm.00$\\
2000 & $.02\pm.00$ & $.03\pm.00$ & $.07\pm.00$ & $.14\pm.00$\\
\hline
\end{tabular}
\end{center}
\caption{Convergence of RMSE for weighted estimator using our weights and constraining $W \in n \Delta^n$, with \textbf{exp} link}
\label{tab:exp-rmse-simplex}

\begin{center}
\footnotesize
\begin{tabular}{cccccccc}
\hline
n & $\textbf{IPS}$ & $\textbf{OptX}_{0.001}$ & $\textbf{OptX}_{0.2}$ & $\textbf{OptX}_{1.0}$ & $\textbf{OptX}_{5.0}$ & $\textbf{DirX}$ & $\textbf{DirZ}$\\
\hline
200 & $.12\pm.01$ & $.76\pm.02$ & $.81\pm.02$ & $.92\pm.02$ & $1.0\pm.02$ & $.10\pm.01$ & $1.0\pm.01$\\
500 & $.11\pm.00$ & $.76\pm.01$ & $.82\pm.01$ & $.92\pm.01$ & $1.0\pm.01$ & $.10\pm.01$ & $1.0\pm.01$\\
1000 & $.10\pm.00$ & $.74\pm.01$ & $.79\pm.01$ & $.90\pm.01$ & $1.0\pm.01$ & $.09\pm.01$ & $1.1\pm.01$\\
2000 & $.10\pm.00$ & $.73\pm.00$ & $.78\pm.00$ & $.88\pm.00$ & $.99\pm.01$ & $.10\pm.01$ & $1.0\pm.01$\\
\hline
\end{tabular}
\end{center}
\caption{Convergence of RMSE for benchmark methods, with \textbf{exp} link}
\label{tab:exp-rmse-bench}

\end{table}

\begin{table}
\begin{center}
\footnotesize
\begin{tabular}{ccccc}
\hline
n & $\textbf{OptZ}_{0.001}$ & $\textbf{OptZ}_{0.2}$ & $\textbf{OptZ}_{1.0}$ & $\textbf{OptZ}_{5.0}$\\
\hline
200 & $.01\pm.06$ & $.00\pm.04$ & $-0.00\pm.04$ & $-0.05\pm.04$\\
500 & $.01\pm.04$ & $.01\pm.03$ & $.00\pm.03$ & $-0.03\pm.03$\\
1000 & $.00\pm.02$ & $.01\pm.02$ & $-0.00\pm.02$ & $-0.01\pm.02$\\
2000 & $.01\pm.01$ & $.01\pm.01$ & $.00\pm.01$ & $.00\pm.01$\\
\hline
\end{tabular}
\end{center}
\caption{Convergence of bias for weighted estimator using our weights, with \textbf{exp} link}
\label{tab:exp-bias}

\begin{center}
\footnotesize
\begin{tabular}{ccccc}
\hline
n & $\textbf{DirX:OptZ}_{0.001}$ & $\textbf{DirX:OptZ}_{0.2}$ & $\textbf{DirX:OptZ}_{1.0}$ & $\textbf{DirX:OptZ}_{5.0}$\\
\hline
200 & $.07\pm.10$ & $.08\pm.06$ & $.10\pm.05$ & $.11\pm.04$\\
500 & $.07\pm.07$ & $.07\pm.05$ & $.09\pm.05$ & $.11\pm.04$\\
1000 & $.06\pm.05$ & $.07\pm.03$ & $.07\pm.03$ & $.09\pm.02$\\
2000 & $.06\pm.04$ & $.06\pm.03$ & $.07\pm.03$ & $.09\pm.03$\\
\hline
\end{tabular}
\end{center}
\caption{Convergence of bias for doubly robust estimator using our weights and $\textbf{DirX}$, with \textbf{exp} link}
\label{tab:exp-bias-drx}

\begin{center}
\footnotesize
\begin{tabular}{ccccc}
\hline
n & $\textbf{DirZ:OptZ}_{0.001}$ & $\textbf{DirZ:OptZ}_{0.2}$ & $\textbf{DirZ:OptZ}_{1.0}$ & $\textbf{DirZ:OptZ}_{5.0}$\\
\hline
200 & $.05\pm.14$ & $.12\pm.10$ & $.23\pm.09$ & $.43\pm.09$\\
500 & $.03\pm.10$ & $.09\pm.07$ & $.16\pm.07$ & $.32\pm.07$\\
1000 & $.02\pm.06$ & $.06\pm.05$ & $.10\pm.05$ & $.23\pm.06$\\
2000 & $.01\pm.04$ & $.03\pm.03$ & $.07\pm.03$ & $.16\pm.04$\\
\hline
\end{tabular}
\end{center}
\caption{Convergence of bias for doubly robust estimator using our weights and $\textbf{DirZ}$, with \textbf{exp} link}
\label{tab:exp-bias-drz}

\begin{center}
\footnotesize
\begin{tabular}{ccccc}
\hline
n & $\textbf{SimplexOptZ}_{0.001}$ & $\textbf{SimplexOptZ}_{0.2}$ & $\textbf{SimplexOptZ}_{1.0}$ & $\textbf{SimplexOptZ}_{5.0}$\\
\hline
200 & $.02\pm.05$ & $.10\pm.06$ & $.20\pm.08$ & $.38\pm.11$\\
500 & $.02\pm.04$ & $.07\pm.04$ & $.14\pm.05$ & $.27\pm.06$\\
1000 & $.01\pm.02$ & $.04\pm.02$ & $.08\pm.03$ & $.19\pm.04$\\
2000 & $.01\pm.01$ & $.03\pm.02$ & $.06\pm.02$ & $.14\pm.02$\\
\hline
\end{tabular}
\end{center}
\caption{Convergence of bias for weighted estimator using our weights and constraining $W \in n \Delta^n$, with \textbf{exp} link}
\label{tab:exp-bias-simplex}

\begin{center}
\footnotesize
\begin{tabular}{cccccccc}
\hline
n & $\textbf{IPS}$ & $\textbf{OptX}_{0.001}$ & $\textbf{OptX}_{0.2}$ & $\textbf{OptX}_{1.0}$ & $\textbf{OptX}_{5.0}$ & $\textbf{DirX}$ & $\textbf{DirZ}$\\
\hline
200 & $.10\pm.06$ & $.74\pm.13$ & $.80\pm.13$ & $.91\pm.15$ & $1.0\pm.16$ & $.07\pm.07$ & $1.0\pm.08$\\
500 & $.11\pm.03$ & $.76\pm.07$ & $.81\pm.08$ & $.92\pm.09$ & $1.0\pm.10$ & $.07\pm.08$ & $1.0\pm.09$\\
1000 & $.10\pm.03$ & $.74\pm.05$ & $.79\pm.05$ & $.90\pm.06$ & $1.0\pm.06$ & $.06\pm.07$ & $1.0\pm.10$\\
2000 & $.10\pm.02$ & $.73\pm.03$ & $.78\pm.03$ & $.88\pm.04$ & $.99\pm.04$ & $.07\pm.07$ & $1.0\pm.09$\\
\hline
\end{tabular}
\end{center}
\caption{Convergence of bias for benchmark methods, with \textbf{exp} link}
\label{tab:exp-bias-bench}

\end{table}

\begin{table}
\begin{center}
\footnotesize
\begin{tabular}{ccccc}
\hline
n & $\textbf{OptZ}_{0.001}$ & $\textbf{OptZ}_{0.2}$ & $\textbf{OptZ}_{1.0}$ & $\textbf{OptZ}_{5.0}$\\
\hline
200 & $.47\pm.04$ & $.35\pm.02$ & $.39\pm.02$ & $.48\pm.02$\\
500 & $.36\pm.05$ & $.27\pm.02$ & $.30\pm.02$ & $.40\pm.02$\\
1000 & $.25\pm.02$ & $.22\pm.01$ & $.25\pm.01$ & $.37\pm.01$\\
2000 & $.14\pm.01$ & $.14\pm.01$ & $.17\pm.01$ & $.27\pm.01$\\
\hline
\end{tabular}
\end{center}
\caption{Convergence of RMSE for weighted estimator using our weights, with \textbf{cubic} link}
\label{tab:cubic-rmse}

\begin{center}
\footnotesize
\begin{tabular}{ccccc}
\hline
n & $\textbf{DirX:OptZ}_{0.001}$ & $\textbf{DirX:OptZ}_{0.2}$ & $\textbf{DirX:OptZ}_{1.0}$ & $\textbf{DirX:OptZ}_{5.0}$\\
\hline
200 & $.58\pm.07$ & $.41\pm.03$ & $.42\pm.02$ & $.38\pm.02$\\
500 & $.37\pm.04$ & $.33\pm.02$ & $.35\pm.02$ & $.37\pm.02$\\
1000 & $.31\pm.02$ & $.31\pm.02$ & $.33\pm.02$ & $.39\pm.01$\\
2000 & $.21\pm.02$ & $.23\pm.02$ & $.26\pm.01$ & $.32\pm.01$\\
\hline
\end{tabular}
\end{center}
\caption{Convergence of RMSE for doubly robust estimator using our weights and $\textbf{DirX}$, with \textbf{cubic} link}
\label{tab:cubic-rmse-drx}

\begin{center}
\footnotesize
\begin{tabular}{ccccc}
\hline
n & $\textbf{DirZ:OptZ}_{0.001}$ & $\textbf{DirZ:OptZ}_{0.2}$ & $\textbf{DirZ:OptZ}_{1.0}$ & $\textbf{DirZ:OptZ}_{5.0}$\\
\hline
200 & $.49\pm.04$ & $.42\pm.03$ & $.54\pm.03$ & $.76\pm.02$\\
500 & $.38\pm.05$ & $.30\pm.02$ & $.38\pm.02$ & $.59\pm.02$\\
1000 & $.27\pm.02$ & $.25\pm.02$ & $.32\pm.02$ & $.52\pm.02$\\
2000 & $.16\pm.01$ & $.16\pm.01$ & $.22\pm.01$ & $.39\pm.01$\\
\hline
\end{tabular}
\end{center}
\caption{Convergence of RMSE for doubly robust estimator using our weights and $\textbf{DirZ}$, with \textbf{cubic} link}
\label{tab:cubic-rmse-drz}

\begin{center}
\footnotesize
\begin{tabular}{ccccc}
\hline
n & $\textbf{SimplexOptZ}_{0.001}$ & $\textbf{SimplexOptZ}_{0.2}$ & $\textbf{SimplexOptZ}_{1.0}$ & $\textbf{SimplexOptZ}_{5.0}$\\
\hline
200 & $.45\pm.04$ & $.41\pm.03$ & $.52\pm.03$ & $.70\pm.03$\\
500 & $.37\pm.05$ & $.30\pm.02$ & $.37\pm.02$ & $.55\pm.02$\\
1000 & $.26\pm.02$ & $.24\pm.02$ & $.31\pm.02$ & $.50\pm.02$\\
2000 & $.14\pm.01$ & $.15\pm.01$ & $.20\pm.01$ & $.35\pm.01$\\
\hline
\end{tabular}
\end{center}
\caption{Convergence of RMSE for weighted estimator using our weights and constraining $W \in n \Delta^n$, with \textbf{cubic} link}
\label{tab:cubic-rmse-simplex}

\begin{center}
\footnotesize
\begin{tabular}{cccccccc}
\hline
n & $\textbf{IPS}$ & $\textbf{OptX}_{0.001}$ & $\textbf{OptX}_{0.2}$ & $\textbf{OptX}_{1.0}$ & $\textbf{OptX}_{5.0}$ & $\textbf{DirX}$ & $\textbf{DirZ}$\\
\hline
200 & $.46\pm.04$ & $1.1\pm.02$ & $1.1\pm.03$ & $1.3\pm.03$ & $1.4\pm.03$ & $.36\pm.02$ & $1.4\pm.01$\\
500 & $.38\pm.02$ & $1.1\pm.01$ & $1.2\pm.01$ & $1.3\pm.01$ & $1.4\pm.01$ & $.34\pm.02$ & $1.4\pm.01$\\
1000 & $.39\pm.01$ & $1.1\pm.01$ & $1.2\pm.01$ & $1.3\pm.01$ & $1.4\pm.01$ & $.35\pm.02$ & $1.4\pm.01$\\
2000 & $.35\pm.01$ & $1.1\pm.01$ & $1.2\pm.01$ & $1.3\pm.01$ & $1.4\pm.01$ & $.39\pm.02$ & $1.4\pm.01$\\
\hline
\end{tabular}
\end{center}
\caption{Convergence of RMSE for benchmark methods, with \textbf{cubic} link}
\label{tab:cubic-rmse-bench}

\end{table}

\begin{table}
\begin{center}
\footnotesize
\begin{tabular}{ccccc}
\hline
n & $\textbf{OptZ}_{0.001}$ & $\textbf{OptZ}_{0.2}$ & $\textbf{OptZ}_{1.0}$ & $\textbf{OptZ}_{5.0}$\\
\hline
200 & $.01\pm.47$ & $.16\pm.31$ & $.29\pm.27$ & $.45\pm.16$\\
500 & $-0.01\pm.36$ & $.12\pm.24$ & $.22\pm.21$ & $.37\pm.15$\\
1000 & $.03\pm.25$ & $.12\pm.18$ & $.20\pm.15$ & $.35\pm.12$\\
2000 & $.02\pm.14$ & $.07\pm.12$ & $.13\pm.11$ & $.26\pm.08$\\
\hline
\end{tabular}
\end{center}
\caption{Convergence of bias for weighted estimator using our weights, with \textbf{cubic} link}
\label{tab:cubic-bias}

\begin{center}
\footnotesize
\begin{tabular}{ccccc}
\hline
n & $\textbf{DirX:OptZ}_{0.001}$ & $\textbf{DirX:OptZ}_{0.2}$ & $\textbf{DirX:OptZ}_{1.0}$ & $\textbf{DirX:OptZ}_{5.0}$\\
\hline
200 & $.13\pm.57$ & $.26\pm.32$ & $.33\pm.27$ & $.34\pm.17$\\
500 & $.12\pm.35$ & $.22\pm.25$ & $.28\pm.21$ & $.34\pm.15$\\
1000 & $.18\pm.26$ & $.24\pm.19$ & $.29\pm.15$ & $.37\pm.11$\\
2000 & $.14\pm.16$ & $.18\pm.14$ & $.22\pm.12$ & $.31\pm.09$\\
\hline
\end{tabular}
\end{center}
\caption{Convergence of bias for doubly robust estimator using our weights and $\textbf{DirX}$, with \textbf{cubic} link}
\label{tab:cubic-bias-drx}

\begin{center}
\footnotesize
\begin{tabular}{ccccc}
\hline
n & $\textbf{DirZ:OptZ}_{0.001}$ & $\textbf{DirZ:OptZ}_{0.2}$ & $\textbf{DirZ:OptZ}_{1.0}$ & $\textbf{DirZ:OptZ}_{5.0}$\\
\hline
200 & $.04\pm.49$ & $.25\pm.34$ & $.45\pm.29$ & $.74\pm.20$\\
500 & $.03\pm.38$ & $.17\pm.25$ & $.32\pm.21$ & $.57\pm.16$\\
1000 & $.04\pm.26$ & $.16\pm.19$ & $.28\pm.16$ & $.50\pm.13$\\
2000 & $.03\pm.16$ & $.10\pm.13$ & $.19\pm.12$ & $.37\pm.10$\\
\hline
\end{tabular}
\end{center}
\caption{Convergence of bias for doubly robust estimator using our weights and $\textbf{DirZ}$, with \textbf{cubic} link}
\label{tab:cubic-bias-drz}

\begin{center}
\footnotesize
\begin{tabular}{ccccc}
\hline
n & $\textbf{SimplexOptZ}_{0.001}$ & $\textbf{SimplexOptZ}_{0.2}$ & $\textbf{SimplexOptZ}_{1.0}$ & $\textbf{SimplexOptZ}_{5.0}$\\
\hline
200 & $.02\pm.45$ & $.22\pm.35$ & $.40\pm.34$ & $.65\pm.28$\\
500 & $.01\pm.37$ & $.15\pm.26$ & $.29\pm.23$ & $.52\pm.18$\\
1000 & $.04\pm.25$ & $.15\pm.19$ & $.27\pm.17$ & $.47\pm.14$\\
2000 & $.02\pm.14$ & $.08\pm.13$ & $.17\pm.11$ & $.34\pm.09$\\
\hline
\end{tabular}
\end{center}
\caption{Convergence of bias for weighted estimator using our weights and constraining $W \in n \Delta^n$, with \textbf{cubic} link}
\label{tab:cubic-bias-simplex}

\begin{center}
\footnotesize
\begin{tabular}{cccccccc}
\hline
n & $\textbf{IPS}$ & $\textbf{OptX}_{0.001}$ & $\textbf{OptX}_{0.2}$ & $\textbf{OptX}_{1.0}$ & $\textbf{OptX}_{5.0}$ & $\textbf{DirX}$ & $\textbf{DirZ}$\\
\hline
200 & $.26\pm.38$ & $1.1\pm.20$ & $1.1\pm.20$ & $1.2\pm.21$ & $1.4\pm.22$ & $.32\pm.16$ & $1.4\pm.12$\\
500 & $.31\pm.22$ & $1.1\pm.10$ & $1.2\pm.10$ & $1.3\pm.10$ & $1.4\pm.11$ & $.29\pm.18$ & $1.4\pm.10$\\
1000 & $.37\pm.14$ & $1.1\pm.07$ & $1.2\pm.07$ & $1.3\pm.08$ & $1.4\pm.08$ & $.32\pm.16$ & $1.4\pm.10$\\
2000 & $.34\pm.09$ & $1.1\pm.05$ & $1.2\pm.06$ & $1.3\pm.06$ & $1.4\pm.06$ & $.34\pm.18$ & $1.4\pm.11$\\
\hline
\end{tabular}
\end{center}
\caption{Convergence of bias for benchmark methods, with \textbf{cubic} link}
\label{tab:cubic-bias-bench}

\end{table}

\begin{table}
\begin{center}
\footnotesize
\begin{tabular}{ccccc}
\hline
n & $\textbf{OptZ}_{0.001}$ & $\textbf{OptZ}_{0.2}$ & $\textbf{OptZ}_{1.0}$ & $\textbf{OptZ}_{5.0}$\\
\hline
200 & $.09\pm.01$ & $.08\pm.01$ & $.13\pm.01$ & $.23\pm.01$\\
500 & $.06\pm.01$ & $.06\pm.00$ & $.09\pm.00$ & $.16\pm.00$\\
1000 & $.04\pm.01$ & $.04\pm.00$ & $.06\pm.00$ & $.12\pm.00$\\
2000 & $.02\pm.00$ & $.03\pm.00$ & $.04\pm.00$ & $.08\pm.00$\\
\hline
\end{tabular}
\end{center}
\caption{Convergence of RMSE for weighted estimator using our weights, with \textbf{linear} link}
\label{tab:linear-rmse}

\begin{center}
\footnotesize
\begin{tabular}{ccccc}
\hline
n & $\textbf{DirX:OptZ}_{0.001}$ & $\textbf{DirX:OptZ}_{0.2}$ & $\textbf{DirX:OptZ}_{1.0}$ & $\textbf{DirX:OptZ}_{5.0}$\\
\hline
200 & $.15\pm.01$ & $.13\pm.01$ & $.13\pm.01$ & $.13\pm.01$\\
500 & $.14\pm.01$ & $.13\pm.01$ & $.13\pm.01$ & $.13\pm.00$\\
1000 & $.13\pm.01$ & $.13\pm.00$ & $.13\pm.00$ & $.13\pm.00$\\
2000 & $.12\pm.00$ & $.12\pm.00$ & $.12\pm.00$ & $.12\pm.00$\\
\hline
\end{tabular}
\end{center}
\caption{Convergence of RMSE for doubly robust estimator using our weights and $\textbf{DirX}$, with \textbf{linear} link}
\label{tab:linear-rmse-drx}

\begin{center}
\footnotesize
\begin{tabular}{ccccc}
\hline
n & $\textbf{DirZ:OptZ}_{0.001}$ & $\textbf{DirZ:OptZ}_{0.2}$ & $\textbf{DirZ:OptZ}_{1.0}$ & $\textbf{DirZ:OptZ}_{5.0}$\\
\hline
200 & $.11\pm.01$ & $.11\pm.01$ & $.18\pm.01$ & $.35\pm.01$\\
500 & $.06\pm.01$ & $.08\pm.01$ & $.13\pm.01$ & $.24\pm.01$\\
1000 & $.05\pm.01$ & $.06\pm.00$ & $.09\pm.00$ & $.18\pm.00$\\
2000 & $.03\pm.00$ & $.04\pm.00$ & $.06\pm.00$ & $.12\pm.00$\\
\hline
\end{tabular}
\end{center}
\caption{Convergence of RMSE for doubly robust estimator using our weights and $\textbf{DirZ}$, with \textbf{linear} link}
\label{tab:linear-rmse-drz}

\begin{center}
\footnotesize
\begin{tabular}{ccccc}
\hline
n & $\textbf{SimplexOptZ}_{0.001}$ & $\textbf{SimplexOptZ}_{0.2}$ & $\textbf{SimplexOptZ}_{1.0}$ & $\textbf{SimplexOptZ}_{5.0}$\\
\hline
200 & $.09\pm.01$ & $.09\pm.01$ & $.15\pm.01$ & $.29\pm.01$\\
500 & $.06\pm.01$ & $.07\pm.01$ & $.10\pm.01$ & $.19\pm.01$\\
1000 & $.04\pm.01$ & $.04\pm.00$ & $.07\pm.00$ & $.14\pm.00$\\
2000 & $.02\pm.00$ & $.03\pm.00$ & $.05\pm.00$ & $.09\pm.00$\\
\hline
\end{tabular}
\end{center}
\caption{Convergence of RMSE for weighted estimator using our weights and constraining $W \in n \Delta^n$, with \textbf{linear} link}
\label{tab:linear-rmse-simplex}

\begin{center}
\footnotesize
\begin{tabular}{cccccccc}
\hline
n & $\textbf{IPS}$ & $\textbf{OptX}_{0.001}$ & $\textbf{OptX}_{0.2}$ & $\textbf{OptX}_{1.0}$ & $\textbf{OptX}_{5.0}$ & $\textbf{DirX}$ & $\textbf{DirZ}$\\
\hline
200 & $.15\pm.01$ & $.57\pm.01$ & $.60\pm.01$ & $.66\pm.01$ & $.72\pm.01$ & $.13\pm.00$ & $.76\pm.00$\\
500 & $.15\pm.01$ & $.57\pm.00$ & $.60\pm.00$ & $.66\pm.00$ & $.72\pm.00$ & $.13\pm.00$ & $.76\pm.00$\\
1000 & $.14\pm.00$ & $.57\pm.00$ & $.60\pm.00$ & $.66\pm.00$ & $.72\pm.00$ & $.13\pm.00$ & $.76\pm.00$\\
2000 & $.14\pm.00$ & $.57\pm.00$ & $.60\pm.00$ & $.66\pm.00$ & $.72\pm.00$ & $.13\pm.00$ & $.76\pm.00$\\
\hline
\end{tabular}
\end{center}
\caption{Convergence of RMSE for benchmark methods, with \textbf{linear} link}
\label{tab:linear-rmse-bench}

\end{table}

\begin{table}
\begin{center}
\footnotesize
\begin{tabular}{ccccc}
\hline
n & $\textbf{OptZ}_{0.001}$ & $\textbf{OptZ}_{0.2}$ & $\textbf{OptZ}_{1.0}$ & $\textbf{OptZ}_{5.0}$\\
\hline
200 & $.03\pm.09$ & $.06\pm.06$ & $.11\pm.05$ & $.23\pm.04$\\
500 & $.02\pm.05$ & $.04\pm.05$ & $.08\pm.04$ & $.15\pm.04$\\
1000 & $.01\pm.04$ & $.03\pm.03$ & $.05\pm.03$ & $.11\pm.03$\\
2000 & $.01\pm.02$ & $.02\pm.02$ & $.04\pm.02$ & $.08\pm.02$\\
\hline
\end{tabular}
\end{center}
\caption{Convergence of bias for weighted estimator using our weights, with \textbf{linear} link}
\label{tab:linear-bias}

\begin{center}
\footnotesize
\begin{tabular}{ccccc}
\hline
n & $\textbf{DirX:OptZ}_{0.001}$ & $\textbf{DirX:OptZ}_{0.2}$ & $\textbf{DirX:OptZ}_{1.0}$ & $\textbf{DirX:OptZ}_{5.0}$\\
\hline
200 & $.12\pm.09$ & $.12\pm.06$ & $.12\pm.05$ & $.13\pm.04$\\
500 & $.13\pm.06$ & $.12\pm.05$ & $.12\pm.04$ & $.12\pm.04$\\
1000 & $.12\pm.04$ & $.12\pm.03$ & $.12\pm.03$ & $.12\pm.03$\\
2000 & $.12\pm.03$ & $.12\pm.03$ & $.12\pm.02$ & $.12\pm.02$\\
\hline
\end{tabular}
\end{center}
\caption{Convergence of bias for doubly robust estimator using our weights and $\textbf{DirX}$, with \textbf{linear} link}
\label{tab:linear-bias-drx}

\begin{center}
\footnotesize
\begin{tabular}{ccccc}
\hline
n & $\textbf{DirZ:OptZ}_{0.001}$ & $\textbf{DirZ:OptZ}_{0.2}$ & $\textbf{DirZ:OptZ}_{1.0}$ & $\textbf{DirZ:OptZ}_{5.0}$\\
\hline
200 & $.04\pm.10$ & $.09\pm.07$ & $.17\pm.06$ & $.34\pm.05$\\
500 & $.02\pm.06$ & $.06\pm.05$ & $.12\pm.05$ & $.24\pm.05$\\
1000 & $.02\pm.04$ & $.05\pm.04$ & $.08\pm.04$ & $.18\pm.04$\\
2000 & $.01\pm.03$ & $.03\pm.03$ & $.06\pm.03$ & $.12\pm.02$\\
\hline
\end{tabular}
\end{center}
\caption{Convergence of bias for doubly robust estimator using our weights and $\textbf{DirZ}$, with \textbf{linear} link}
\label{tab:linear-bias-drz}

\begin{center}
\footnotesize
\begin{tabular}{ccccc}
\hline
n & $\textbf{SimplexOptZ}_{0.001}$ & $\textbf{SimplexOptZ}_{0.2}$ & $\textbf{SimplexOptZ}_{1.0}$ & $\textbf{SimplexOptZ}_{5.0}$\\
\hline
200 & $.02\pm.08$ & $.07\pm.06$ & $.14\pm.06$ & $.29\pm.06$\\
500 & $.02\pm.05$ & $.05\pm.05$ & $.09\pm.05$ & $.19\pm.04$\\
1000 & $.01\pm.04$ & $.03\pm.03$ & $.06\pm.03$ & $.14\pm.03$\\
2000 & $.01\pm.02$ & $.02\pm.02$ & $.04\pm.02$ & $.09\pm.02$\\
\hline
\end{tabular}
\end{center}
\caption{Convergence of bias for weighted estimator using our weights and constraining $W \in n \Delta^n$, with \textbf{linear} link}
\label{tab:linear-bias-simplex}

\begin{center}
\footnotesize
\begin{tabular}{cccccccc}
\hline
n & $\textbf{IPS}$ & $\textbf{OptX}_{0.001}$ & $\textbf{OptX}_{0.2}$ & $\textbf{OptX}_{1.0}$ & $\textbf{OptX}_{5.0}$ & $\textbf{DirX}$ & $\textbf{DirZ}$\\
\hline
200 & $.13\pm.08$ & $.57\pm.05$ & $.60\pm.05$ & $.66\pm.05$ & $.72\pm.05$ & $.13\pm.04$ & $.76\pm.04$\\
500 & $.14\pm.05$ & $.57\pm.04$ & $.60\pm.03$ & $.66\pm.03$ & $.72\pm.03$ & $.12\pm.04$ & $.76\pm.02$\\
1000 & $.14\pm.04$ & $.57\pm.02$ & $.60\pm.02$ & $.66\pm.02$ & $.72\pm.02$ & $.13\pm.03$ & $.76\pm.03$\\
2000 & $.13\pm.03$ & $.57\pm.02$ & $.60\pm.02$ & $.66\pm.02$ & $.72\pm.02$ & $.13\pm.03$ & $.76\pm.02$\\
\hline
\end{tabular}
\end{center}
\caption{Convergence of bias for benchmark methods, with \textbf{linear} link}
\label{tab:linear-bias-bench}

\end{table}

\end{document}